\documentclass{article}

\usepackage[utf8]{inputenc}
\usepackage[T1]{fontenc}

\usepackage{microtype}
\usepackage{graphicx}
\usepackage{subcaption}
\usepackage{booktabs} 
\usepackage{multirow} 
\usepackage{float} 

\usepackage{hyperref}
\usepackage{url}

\usepackage{PRIMEarxiv}

\usepackage{natbib}

\usepackage{amsmath}
\usepackage{amsfonts}
\usepackage{amssymb}
\usepackage{mathtools}
\usepackage{amsthm}

\usepackage[capitalize,noabbrev]{cleveref}

\theoremstyle{plain}

\theoremstyle{definition}

\theoremstyle{remark}

\usepackage[disable,textsize=tiny]{todonotes}

\usepackage{xcolor}
\usepackage{colortbl} 
\definecolor{backcolor}{gray}{0.92} 

\definecolor{promptbg}{gray}{0.95}
\definecolor{promptframe}{gray}{0.4}

\newsavebox{\promptboxcontent}
\newenvironment{promptbox}[1]{%
  \begin{lrbox}{\promptboxcontent}%
  \begin{minipage}{\dimexpr\linewidth-16pt\relax}%
  \textbf{#1}\par\smallskip
}{%
  \end{minipage}%
  \end{lrbox}%
  \noindent\fcolorbox{promptframe}{promptbg}{\usebox{\promptboxcontent}}%
}

\newcommand{\name}{\textsc{ToolSelf}}

\title{\name{}: Unifying Task Execution and Self-Reconfiguration via Tool-Driven Emergent Adaptation}

\author{
  Jingqi Zhou$^{1}$, Sheng Wang$^{1}$, DeZhao Deng$^{1}$, Junwen Lu$^{1}$, Junwei Su$^{1}$, \\
  Qintong Li$^{1}$, Jiahui Gao$^{1}$, Hao Wu$^{1}$, Jiyue Jiang$^{2}$, Lingpeng Kong$^{1}$, Dunhong Jin$^{1}$, Chuan Wu$^{1}$ \\
  $^{1}$The University of Hong Kong \\
  $^{2}$The Chinese University of Hong Kong \\
  \texttt{\{u3011211, u3009618, u3638316, u3664724\}@connect.hku.hk} \\
  \texttt{\{junweisu, qtli, sumiler\}@connect.hku.hk, wuhao55@hku.hk} \\
  \texttt{jiangjy@link.cuhk.edu.hk, \{lpk, dhjin, cwu\}@cs.hku.hk} \\
}

\begin{document}

\maketitle

\begin{abstract}
LLM-powered agentic systems excel at complex long-horizon tasks, but remain constrained by static configurations fixed before execution. Such rigidity forces a trade-off between domain-specific performance and cross-task generalization: strong priors and compact tool spaces aid specialization but weaken transfer, while task-agnostic workflows and broad action spaces expand coverage but dilute guidance. Existing pre-execution optimization, planner-worker orchestration, and configuration patching fall short of resolving this tension, as they decouple adaptation from execution, causing information loss, fragmented optimization, and ambiguous credit assignment. We propose \name{}, a \textbf{tool}-driven runtime \textbf{self}-reconfiguration paradigm that abstracts configuration updates as a standardized tool interface and unifies execution and adaptation within one policy's action space. The execution agent can dynamically update sub-goals, strategies, toolboxes, context, and context-management modes based on task progress and feedback. We further introduce \textbf{C}onfiguration-\textbf{A}ware Two-stage \textbf{T}raining (CAT), which combines rejection sampling fine-tuning with trajectory-level KTO reinforcement learning to internalize self-reconfiguration. Across diverse benchmarks, zero-shot \name{} rivals task-specialized agents; after CAT training, \name{} gains 28.8 points over the static-configuration baseline on average, illuminating a path toward emergent adaptivity that obviates manually injected guidance.\footnote{The code is available at \url{https://github.com/lian-tian-mo-zun/ToolSelf}.}
\end{abstract}

\section{Introduction}
\label{sec:Introduction}

\begin{figure}[!t]
\centering
\includegraphics[width=\textwidth]{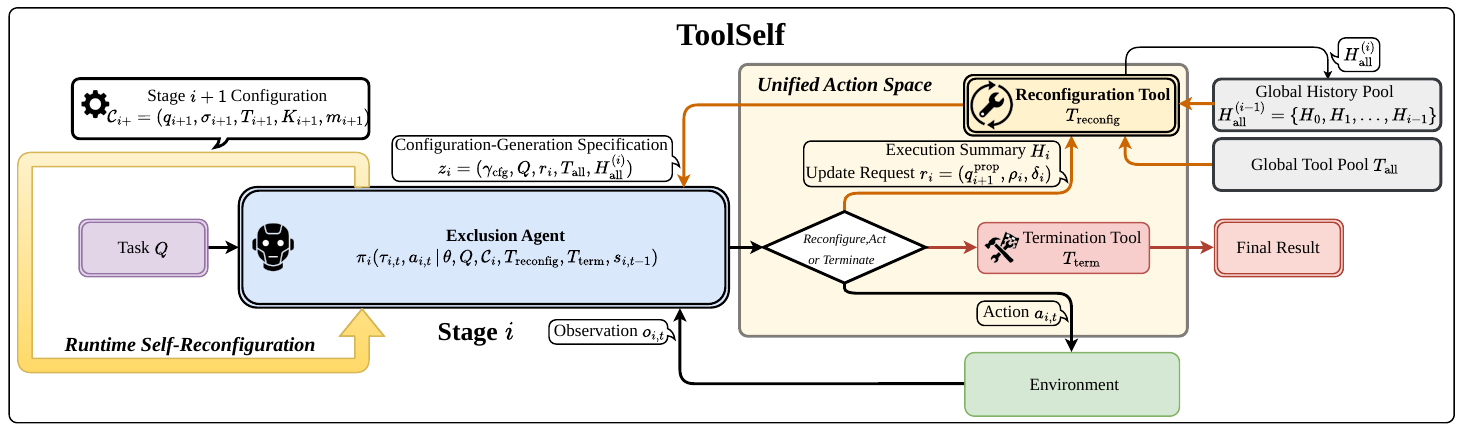}

\caption{Overview of \name{}: Instead of using a static external configuration, \name{} treats configuration as a dynamic, tool-updatable variable. At stage $i$, the execution agent $\pi_i$ operates under $\mathcal{C}_i=(q_i,\sigma_i,T_i,K_i,m_i)$, which defines the current sub-goal, execution strategy, stage-specific toolbox, task knowledge, and context mode. If the configuration misaligns with task progress, $\pi_i$ summarizes the stage and invokes the tool $T_{\text{reconfig}}$, which provides a specification guiding $\pi_i$ to generate the next configuration, $\mathcal{C}_{i+1}$. Ultimately, the core innovation of \name{} lies in unifying dynamic runtime adaptation and task execution within a single, end-to-end optimizable policy, utilizing a standardized execution-reconfiguration loop to foster emergent adaptability instead of relying on manually injected guidance.}
\label{fig:main}

\end{figure}

\emph{Agentic systems} refer to large language model--based systems that go beyond passive text generation to autonomously plan, reason, and act in interactive environments through multi-step decision making~\citep{kolt2025governing}. By tightly coupling internal reasoning with external actions, such as tool invocation, environment interaction, and intermediate state updates, agentic systems enable models to tackle complex long-horizon tasks that cannot be solved through single-shot inference~\citep{yao2022react}. This paradigm has become increasingly important as real-world applications, including deep research~\citep{wu2025agentic}, software engineering~\citep{yang2024swe}, and system control~\citep{wang2024appbench, qin2025ui}, demand sustained goal pursuit, adaptive planning, and iterative integration of information over time.

Central to the behavior of an agentic system is its \emph{configuration}, which defines the internal control variables governing how the agent reasons, acts, and interacts with its environment during execution. Configuration typically includes the agent's execution strategy (e.g., persona or reasoning style)~\citep{chen2023autoagents}, the available action toolbox~\citep{lu2025octotools}, and cross-step context and its management~\citep{zhou2025mem1}. These elements implicitly shape the agent's action space, information flow, and decision-making dynamics, and therefore play a decisive role in long-horizon performance~\citep{guo2024large}.
However, most existing agentic systems adopt a \emph{static configuration} fixed before execution and unchanged throughout the task.
This rigidity induces a fundamental tension between domain-specific performance and generality~\citep{xia2024agentless}. Achieving high performance typically requires encoding domain priors into the configuration to strengthen task guidance~\citep{yang2024swe, antoniades2024swe, li2025websailor, li2025search, zhang2025cosight}; yet this ties the agent to specific tasks and weakens transferability. Conversely, a general-purpose agent must rely on task-agnostic workflows, prompts, context strategies, and broad toolsets~\citep{yao2022react, wang2025openhands}; although this broadens task coverage, it also expands the action space, dilutes task priors, and ultimately degrades long-horizon performance~\citep{xia2024agentless}.

\paragraph{Limitation of Existing Approaches.}
Prior work (Appendix~\ref{sec:related_work}) has mainly pursued three directions. \emph{Pre-execution optimization} optimizes agents offline before execution to automatically derive task-specialized configurations, but the resulting configurations remain static and cannot adapt at runtime~\citep{zhang2024aflow, shang2024agentsquare, tang2025autoagent}. \emph{Planner-worker orchestration} uses a centralized planner to coordinate specialized workers for heterogeneous tasks, improving generality while retaining domain performance; however, planner-mediated communication makes task states vulnerable to information loss and errors under repeated relay and compression~\citep{hu2025owl, fourney2024magentic}. \emph{Configuration patching} uses external strategies to adapt individual configuration dimensions of a general agent, such as pruning toolsets~\citep{betser2026agentrim}, managing context~\citep{wu2025resum, zhou2025mem1}, or strengthening prompt guidance~\citep{zhu2025oagents, sun2023adaplanner}; yet such adaptation remains fragmented and rarely yields consistent system-level gains.
Moreover, these methods generally decouple task adaptation from execution policy: external optimizers, planners, or auxiliary modules adjust the configuration, while the execution agent advances the task. This separation prevents adaptation policies from perceiving execution details, often yielding configurations misaligned with task demands, and obscures credit assignment between adaptation and execution, thereby hindering joint optimization.

\paragraph{Challenges and Design Requirements.}
The limitations of these methods highlight the key challenge in enabling truly adaptive agents: how to tightly integrate task adaptation and execution at runtime, and represent configuration updates through a \emph{standardized and minimal interface}, thereby replacing externally driven strategies or fragmented optimizations over individual configuration dimensions. Such a unified mechanism can reduce the information loss caused by cross-strategy relay and compression, while placing configuration decisions and execution behavior on the same trajectory, thereby supporting end-to-end joint optimization.

\paragraph{Our Approach.}
To this end, we propose \name{} (\textbf{Tool}-driven \textbf{Self}-reconfiguration), a novel paradigm that abstracts configuration updates as a standardized tool interface and unifies task execution and self-reconfiguration within the single action space of one policy, thereby enabling the joint optimization and emergence of both capabilities. As illustrated in Figure~\ref{fig:main}, unlike conventional architectures that decouple task adaptation from execution, \name{} equips the execution agent with a reconfiguration tool. Conditioned on task progression and environmental feedback, the agent can invoke this tool to directly update its configuration, autonomously and comprehensively adjusting several core configuration dimensions, including sub-goals, execution strategy, toolbox, and context together with its management scheme. This enables runtime adaptation to the specialized requirements of diverse tasks and various execution stages while preserving cross-task generalization.

Moreover, because task execution and configuration adaptation are unified within the same policy and action space, \name{} naturally supports end-to-end joint optimization. To this end, we propose the Configuration-Aware Two-stage Training (CAT) framework: Stage I employs rejection sampling fine-tuning (RFT)~\citep{li2025websailor} for cold-start initialization using successful trajectories generated by teacher models; Stage II uses KTO-based reinforcement learning~\citep{ethayarajh2024kto} to further optimize reconfiguration decisions. Since self-reconfiguration is a cross-stage meta-capability whose quality only manifests through downstream task performance, we adopt trajectory-level credit assignment by propagating task success or failure to all reconfiguration decisions in the trajectory; KTO's binary feedback is naturally aligned with this setting.

Extensive experiments across deep research, general AI assistance, and software engineering benchmarks demonstrate that \name{}, even under zero-shot evaluation, already rivals task-specialized agents while exhibiting strong cross-task generalization (Sec.~\ref{sec:main_results}); after CAT training, \name{} gains 28.8 points over the static-configuration baseline on average (Sec.~\ref{sec:training_results}). Ablation studies and case analyses further validate the effectiveness of each core design (Sec.~\ref{sec:ablation}, Appendix~\ref{sec:case_studies}), showing that \name{} not only overcomes the performance degradation induced by cross-policy information exchange, but also elicits advanced capabilities such as long-horizon planning, self-refinement, and self-correction, thereby illuminating a promising path toward agentic systems with learnable adaptivity.
The main contributions are summarized as follows:
\begin{itemize}
\item We propose \name{}, a paradigm that abstracts configuration updates as a standardized tool interface and unifies task execution and self-reconfiguration within the single action space of one policy, thereby enabling the joint optimization and emergence of both capabilities.
\item We introduce Configuration-Aware Two-stage Training (CAT), an end-to-end framework that first cold-starts \name{} with RFT and then refines cross-stage self-reconfiguration via KTO-based reinforcement learning with trajectory-level binary feedback.
\item Across deep research, general AI assistance, and software engineering benchmarks, zero-shot \name{} rivals task-specialized agents without sacrificing cross-task generalization. After CAT training, \name{} gains 28.8 points over the static-configuration baseline on average and elicits advanced capabilities spanning long-horizon planning, self-refinement, and self-correction, illuminating a path toward emergent adaptivity that obviates manually injected guidance.
\end{itemize}

\section{Method}
\label{sec:Method}

\name{} enables runtime self-reconfiguration in a single agent by \textbf{abstracting configuration updates as a standardized tool interface}, thereby unifying task execution and configuration adaptation within the single action space of one policy.
Formally, this paradigm is characterized by four properties that distinguish it from existing approaches: (1) \textbf{Runtime Specialization}, where the agent dynamically updates its configuration during execution rather than relying on a static configuration derived before execution; (2) \textbf{Integrated Adaptation and Execution}, where a single policy is responsible for both task progress and configuration adaptation, instead of delegating adaptation to a pre-execution optimizer, centralized planner, or external auxiliary module; (3) \textbf{Standardized and Comprehensive Updating}, where the major configuration dimensions are updated through one unified interface rather than through fragmented patches to isolated dimensions; and (4) \textbf{Joint Optimization}, where configuration decisions and task execution reside on the same trajectory and are therefore amenable to end-to-end optimization under unified downstream feedback.
The system operates through an \textit{execution-reconfiguration} loop: the execution agent advances the task under the current configuration and invokes the reconfiguration tool when adaptation is needed; the tool returns a configuration-generation specification that guides the same policy to synthesize an updated configuration and continue execution.

\subsection{Preliminaries: Pre-Fixed Configuration Paradigm}
\label{sec:preliminaries}

Formally, the behavior policy of a ReAct-style~\citep{yao2022react} single agent can be described by the following conditional distribution:
\begin{equation}
\pi\!\left(\tau_t, a_t \,\middle|\, \theta, Q, \mathcal{C}_{\text{static}}, s_{t-1} \right)
\end{equation}
Here, $\theta$ denotes the LLM parameters, $Q$ is the target task, and $\mathcal{C}_{\text{static}} \triangleq (\sigma, T, K, m)$ is the static configuration, consisting of the execution strategy $\sigma$, toolbox $T$, background knowledge $K$, and context-management mode $m$. The agent's raw interaction history is
\[
s_{t-1} = (\tau_0, a_0, o_0, \ldots, \tau_{t-1}, a_{t-1}, o_{t-1}),
\]
where $\tau_k$, $a_k$, and $o_k$ denote the thought, action, and observation at step $k$, respectively. The mode $m$ specifies how $s_{t-1}$ is maintained, compressed, truncated, and organized, thereby producing the visible context $\tilde{s}_{t-1}=M_m(s_{t-1})$ exposed to the agent. Under the pre-fixed configuration paradigm, $m$ is determined before execution and remains unchanged throughout the task.

The fundamental limitation is that $\mathcal{C}_{\text{static}}$ is fixed before the task begins and remains frozen during execution, forcing an unfavorable specialization--generalization trade-off. Strong domain priors provide focused task guidance and a compact tool space, but tie the agent to specific tasks and weaken cross-task generalization; a general configuration broadens task coverage, but exposes the agent to a larger action space, weaker task priors, and more difficult context management. More critically, with configuration variables outside the agent's action space, the executing agent cannot proactively adapt them to stage objectives, tool requirements, knowledge gaps, or context state, and must continue under the initial setup, leading to inefficient exploration, accumulated information pollution, and degraded long-horizon performance.

\subsection{The \name{} Paradigm}

To overcome these limitations, \name{} reconceptualizes configuration as a dynamic, tool-updatable state rather than a pre-fixed constant. As illustrated in Figure~\ref{fig:main}, \name{} equips the execution agent $\pi_i$ with a standardized reconfiguration tool $T_{\text{reconfig}}$, thereby representing configuration adaptation as a native tool call in the same policy's action space. Conditioned on task progression and environmental feedback, $\pi_i$ invokes this tool to submit a stage execution summary and a reconfiguration request; the tool then returns a configuration-generation specification that guides $\pi_i$ to produce the next-stage configuration $\mathcal{C}_{i+1}$.

We first formalize the dynamic configuration at each stage. A stage is defined as the execution interval between two consecutive reconfigurations, with its configuration given by:
\begin{equation}
\mathcal{C}_i \triangleq (q_i, \sigma_i, T_i, K_i, m_i)
\label{eq:dynamic_config}
\end{equation}
Unlike the pre-fixed $\mathcal{C}_{\text{static}} = (\sigma, T, K, m)$ that lies outside the agent's action space, this formulation introduces the sub-goal $q_i$ and allows every configuration element to evolve with the stage index $i$. Specifically, $q_i$ specifies the current-stage objective; $\sigma_i$ denotes the execution strategy; $T_i \subseteq T_{\text{all}}$ is the stage-specific toolbox selected from the global tool pool; $K_i$ is the local knowledge accessible at the current stage, including task-relevant facts, prior progress, and key constraints; and $m_i$ denotes the context-management mode, which determines how the execution agent maintains, compresses, truncates, and organizes the visible context within stage $i$.
The system further maintains a global history sequence $H_{\text{all}}^{(i)}$, which records execution summaries up to stage $i$.

Building on this configuration, we formalize the execution agent and its reconfiguration tool as follows.

\textbf{Execution Agent} operates under the behavior policy:
\begin{equation}
\pi_i\!\left( \tau_{i,t}, a_{i,t} \,\middle|\, \theta, Q, \mathcal{C}_i, T_{\text{reconfig}}, T_{\text{term}}, s_{i,t-1} \right)
\label{eq:inference_policy}
\end{equation}
At stage $i$, the agent operates under configuration $\mathcal{C}_i$ and may invoke $T_{\text{reconfig}}$ to trigger configuration updates or the termination tool $T_{\text{term}}$ to terminate execution. The notation $\pi_i$ denotes the stage-conditioned behavior distribution induced by the same execution policy $\pi_\theta$ under configuration $\mathcal{C}_i$; behavioral differences across stages arise from the dynamic configuration rather than from different model parameters.

\textbf{Reconfiguration tool $T_{\text{reconfig}}$.} When the execution agent $\pi_i$ determines that the current configuration $\mathcal{C}_i$ is no longer aligned with subsequent task requirements, it first summarizes the current stage execution as $H_i$ and generates a reconfiguration request
\begin{equation}
r_i \triangleq (q_{i+1}^{\text{prop}}, \rho_i, \delta_i)
\label{eq:reconfig_request}
\end{equation}
where $q_{i+1}^{\text{prop}}$ denotes the proposed next-stage sub-goal, $\rho_i$ specifies the reconfiguration rationale, and $\delta_i$ describes the required adjustments to the next-stage execution strategy, toolbox, knowledge, and context-management mode.

The agent then invokes $T_{\text{reconfig}}$, which first archives the stage summary into the global history:
\begin{equation}
H_{\text{all}}^{(i)} \leftarrow H_{\text{all}}^{(i-1)} \oplus (H_i).
\end{equation}
It then returns a \textbf{configuration-generation specification} that guides and constrains the generation of the next-stage configuration:
\begin{equation}
z_i = (\gamma_{\text{cfg}}, Q, r_i, T_{\text{all}}, H_{\text{all}}^{(i)}),
\label{eq:reconfig_context}
\end{equation}
where $\gamma_{\text{cfg}}$ is a prompt specifying the required format and constraints for configuration generation. The tool response $z_i$ is appended after $s_{i,t-1}$, guiding $\pi_i$ to generate the next-stage configuration:
\begin{equation}
\pi_i\!\left(
\mathcal{C}_{i+1}
\,\middle|\,
\theta, Q, \mathcal{C}_i, T_{\text{reconfig}}, T_{\text{term}}, s_{i,t-1}, z_i
\right).
\label{eq:reconfig_engine}
\end{equation}

This design divides configuration updates into two phases: intent expression and configuration generation. During task execution, $\pi_i$ is exposed only to the stage toolbox $T_i$, which keeps the action space compact and task-relevant. The full tool pool $T_{\text{all}}$ is exposed through $z_i$ only after the agent explicitly invokes $T_{\text{reconfig}}$, and even then it serves only as the candidate space for constructing the next-stage toolbox $T_{i+1}$ rather than as a set of directly callable tools. This separation prevents configuration-generation instructions from interfering with task execution, reduces erroneous tool invocation, and improves runtime stability.

Based on the above tool-mediated configuration generation, the complete execution process of \name{} can be described as the following \textit{execution-reconfiguration} loop, as shown in Figure~\ref{fig:main}:

\textbf{1. Initialization:} Given the main task $Q$, the system constructs the initial reconfiguration context $z_{-1}=(\gamma_{\text{cfg}},Q,r_{-1},T_{\text{all}},H_{\text{all}}^{(-1)})$, where $r_{-1}=\emptyset$ and $H_{\text{all}}^{(-1)}=\emptyset$. It then invokes the same LLM that drives the execution agent to generate the initial configuration $\mathcal{C}_0=(q_0,\sigma_0,T_0,K_0,m_0)$ based on $z_{-1}$, and starts execution under this configuration.

\textbf{2. Task Execution:} In stage $i$, the execution agent $\pi_i$ pursues the sub-goal $q_i$ under configuration $\mathcal{C}_i$. When $\pi_i$ determines that the current configuration can no longer satisfy task requirements, for example because the sub-goal $q_i$ has been completed, toolbox $T_i$ is insufficient, knowledge $K_i$ is missing, strategy $\sigma_i$ has failed, or context-management mode $m_i$ is no longer appropriate for subsequent execution, it first summarizes the stage execution as $H_i$. It then generates the reconfiguration request $r_i = (q_{i+1}^{\text{prop}}, \rho_i, \delta_i)$ and invokes $T_{\text{reconfig}}(H_i, r_i)$.

\textbf{3. Reconfiguration:} $T_{\text{reconfig}}$ archives $H_i$ into $H_{\text{all}}^{(i)}$ and returns the structured reconfiguration context $z_i=(\gamma_{\text{cfg}},Q,r_i,T_{\text{all}},H_{\text{all}}^{(i)})$. This context guides $\pi_i$ to generate the new configuration $\mathcal{C}_{i+1}$, after which a new execution agent is launched under policy $\pi_{i+1}$. The new agent continues with configuration $\mathcal{C}_{i+1}$: it recovers prior progress and key information from $K_{i+1}$, clarifies the new objective through $q_{i+1}$, and begins with an empty stage-local history $s_{i+1}$.

This loop repeats until the execution agent determines that the main task $Q$ has been fully resolved and invokes $T_{\text{term}}$ to terminate execution.

\subsection{Configuration-Aware Two-stage Training}

Although \name{} can operate without task-specific training, dedicated optimization can further improve its performance. We first formalize the execution process of \name{}. A complete execution trajectory is defined as $\mathcal{T} = ((\mathcal{C}_0, \xi_0), (\mathcal{C}_1, \xi_1), \ldots, (\mathcal{C}_N, \xi_N))$, where $\xi_i = (\tau_{i,0}, a_{i,0}, o_{i,0}, \ldots, \tau_{i,M_i}, a_{i,M_i}, o_{i,M_i})$ denotes the execution trajectory at stage $i$. The generation probability of one stage trajectory is:
\begin{equation}
p(\xi_i \mid \theta, Q, \mathcal{C}_i) = \prod_{t=0}^{M_i} \pi_i(\tau_{i,t}, a_{i,t} \mid s_{i,t-1}) \cdot p(o_{i,t} \mid a_{i,t})
\end{equation}
The joint probability distribution of the complete trajectory factorizes as:
\begin{equation}
\resizebox{0.92\linewidth}{!}{$\displaystyle
p(\mathcal{T}\mid \theta,Q,T_{\text{all}})
= p_\theta(\mathcal{C}_0\mid z_{-1})
\times \prod_{i=0}^{N} p_\theta(\xi_i\mid Q,\mathcal{C}_i)
\times \prod_{i=0}^{N-1} p_\theta(\mathcal{C}_{i+1}\mid Q,\mathcal{C}_i,\xi_i,z_i).
$}
\end{equation}
This factorization provides the foundation for the training framework.

To internalize self-reconfiguration as a meta-capability, we propose \textbf{Configuration-Aware Two-stage Training (CAT)}, which trains a single execution policy $\pi_\theta$ to jointly learn task execution, reconfiguration triggering, stage summarization, reconfiguration-request generation, and next-stage configuration generation. CAT consists of Rejection Sampling Fine-tuning (RFT)~\citep{li2025websailor} for cold-start initialization and Kahneman-Tversky Optimization (KTO)~\citep{ethayarajh2024kto} for reinforcement learning refinement.

\textbf{Stage I: Cold-Start Initialization.} To provide strong priors for configuration decisions, we bootstrap from teacher demonstrations~\citep{li2025websailor}. We use a teacher model to generate trajectories on the task dataset and filter successful ones through rejection sampling. For task $Q^{(j)}$, let $\mathcal{T}_{\text{teacher}}^{(j)}$ denote the successful trajectory obtained by rejection sampling, and define a binary success label $y(\mathcal{T}_{\text{teacher}}^{(j)}) \in \{0,1\}$, where 1 indicates successful task completion and 0 indicates failure. The filtered dataset is defined as:
\begin{equation}
\mathcal{D}_{\text{SFT}}
=
\left\{
(\mathcal{T}_{\text{teacher}}^{(j)}, Q^{(j)})
\right\}_{j=1}^{M}.
\end{equation}
Each $\mathcal{T}_{\text{teacher}}^{(j)}$ is sampled from $p(\mathcal{T}\mid \theta_{\text{teacher}},Q^{(j)})$ and satisfies $y(\mathcal{T}_{\text{teacher}}^{(j)})=1$. Let $N^{(j)}$ denote the final stage index of trajectory $\mathcal{T}_{\text{teacher}}^{(j)}$. For each non-terminal stage, $\xi_i^{(j)}$ ends with a reconfiguration tool call; hence the stage summary $H_i^{(j)}$ and reconfiguration request $r_i^{(j)}$ are supervised as part of the stage trajectory. The policy then generates the next-stage configuration $\mathcal{C}_{i+1}^{(j)}$ conditioned on the tool-returned $z_i^{(j)}$. RFT therefore supervises both stage execution and configuration generation:
\begin{equation}
\resizebox{0.92\linewidth}{!}{$\displaystyle
\mathcal{L}_{\text{RFT}}
= -\sum_{j=1}^{M}\left[
\sum_{i=0}^{N^{(j)}} \log p_\theta(\xi_i^{(j)} \mid Q^{(j)}, \mathcal{C}_i^{(j)})
+
\sum_{i=0}^{N^{(j)}-1} \log p_\theta(\mathcal{C}_{i+1}^{(j)} \mid Q^{(j)}, \mathcal{C}_i^{(j)}, \xi_i^{(j)}, z_i^{(j)})
\right].
$}
\end{equation}
We denote the policy parameters after RFT as $\theta^{\text{RFT}}$.

\textbf{Stage II: Reinforcement Learning Refinement.} To move beyond imitation, we further optimize the same execution policy with KTO, which directly leverages binary feedback. The KTO loss is defined as:
\begin{equation}
\ell_{\text{KTO}}(x,y;\theta,\theta_{\text{ref}})
=
\begin{cases}
\lambda_D \cdot v(r(x)-z_0), & \text{if } y=1, \\
\lambda_U \cdot v(z_0-r(x)), & \text{if } y=0,
\end{cases}
\end{equation}
where $r(x)=\log \frac{p_\theta(x)}{p_{\theta_{\text{ref}}}(x)}$ is the log-probability ratio, $v(\cdot)$ is the value function, $z_0$ is the KL target, and $\lambda_D,\lambda_U$ are weighting coefficients.

Self-reconfiguration is a cross-stage meta-capability whose quality manifests only through downstream execution, making step-level credit assignment impractical. We therefore adopt trajectory-level inheritance: for each sampled trajectory $\mathcal{T}_k^{(j)}$, its binary label $y_k^{(j)}$ is propagated to all stage-execution and configuration-generation decisions along that trajectory. Specifically, for the $M'$ tasks in the training set, we sample $K$ trajectories with the RFT model and assign binary labels $y_k^{(j)}\in\{0,1\}$ according to task success or failure. Using $\theta^{\text{RFT}}$ as the reference model, the KTO objective is:
\begin{equation}
\resizebox{0.92\linewidth}{!}{$\displaystyle
\mathcal{L}_{\text{KTO}}
= \sum_{j=1}^{M'} \sum_{k=1}^{K}\left[
\sum_{i=0}^{N_k^{(j)}} \ell_{\text{KTO}}(\xi_{i,k}^{(j)}, y_k^{(j)}; \theta,\theta^{\text{RFT}})
+
\sum_{i=0}^{N_k^{(j)}-1} \ell_{\text{KTO}}(\mathcal{C}_{i+1,k}^{(j)}, y_k^{(j)}; \theta,\theta^{\text{RFT}})
\right].
$}
\end{equation}
The terminal stage does not include the $\mathcal{C}_{i+1}$ term. Through this unified objective, the agent learns to coordinate task execution and self-reconfiguration within the same policy, so that configuration updates are optimized explicitly for downstream task success.

\subsection{Design Advantages}

\textbf{Runtime Adaptivity.} \name{} turns configuration from a static pre-execution setting into a runtime-updatable variable, enabling stage-wise behavioral adaptation and easing the specialization--generalization trade-off.

\textbf{Reduced Coordination Loss.} By colocating execution and adaptation within a single policy, \name{} conditions configuration updates directly on execution details and environmental feedback, avoiding state relay, compression, and dispatch through external optimizers, planners, or auxiliary modules.

\textbf{Holistic Configuration Control.} Through a unified interface, \name{} jointly updates sub-goals, execution strategies, toolboxes, knowledge, and context-management modes, maintaining coherent stage-local execution states rather than patching isolated configuration dimensions.

\textbf{Trainable and Emergent Self-Reconfiguration.} With configuration decisions and execution behavior embedded in the same trajectory, CAT propagates downstream task feedback to reconfiguration triggering, stage summarization, request generation, and configuration generation, turning self-reconfiguration from a handcrafted patch into an optimizable policy capability that elicits long-horizon planning, self-refinement, and self-correction.

\section{Experiments}
\label{sec:Experiments}

\begin{table}[t]
    \caption{Main results across five evaluation datasets and two model scales. We compare \name{} with baselines from agents with static configuration, pre-execution optimization, planner-worker orchestration, and configuration patching. FRAMES and xbench evaluate deep research, GAIA evaluates general AI assistant tasks, and SWE-Lite denotes SWE-bench Lite. GAIA(WS) is the WebSailor subset of GAIA containing tasks solvable by search and browsing only. Avg.\ denotes the average over available datasets. Bold indicates the strongest result and underline indicates the second strongest result in each column.}
    \label{tab:main_results}
    \centering
    \resizebox{\textwidth}{!}{%
    \begin{tabular}{lcccccccccccc}
    \toprule
    \raisebox{-0.5\normalbaselineskip}[0pt][0pt]{\textbf{Method}} & \multicolumn{6}{c}{\texttt{Qwen3-14B}} & \multicolumn{6}{c}{\texttt{Qwen3-8B}} \\
    \cmidrule(lr){2-7} \cmidrule(lr){8-13}
    & FRAMES & xbench & GAIA(WS) & GAIA & SWE-Lite & Avg. & FRAMES & xbench & GAIA(WS) & GAIA & SWE-Lite & Avg. \\
    \midrule
    \midrule
    \rowcolor{backcolor}
    \multicolumn{13}{c}{\textit{General-Purpose Agents with Static Configuration}} \\
    \midrule
    Vanilla Agent & 38.0 & 13.0 & 33.0 & \underline{32.1} & 13.3 & \underline{25.9} & 21.0 & 6.0 & 19.4 & 19.7 & 10.1 & 15.2 \\
    OpenHands & 36.0 & 11.0 & 30.1 & 28.5 & 14.1 & 23.9 & 24.0 & 7.0 & 18.4 & 18.2 & 10.8 & \underline{15.7} \\
    \midrule
    \rowcolor{backcolor}
    \multicolumn{13}{c}{\textit{Task-Specialized Agents with Static Configuration}} \\
    \midrule
    WebSailor & 42.0 & 15.0 & 35.0 & -- & -- & -- & 28.0 & 4.0 & 18.5 & -- & -- & -- \\
    SWE-Agent & -- & -- & -- & -- & 14.2 & -- & -- & -- & -- & -- & 10.9 & -- \\
    Co-Sight & 46.0 & \underline{16.0} & 31.1 & -- & -- & -- & 32.0 & \underline{13.0} & \underline{28.1} & -- & -- & -- \\
    SWE-Search & -- & -- & -- & -- & \underline{14.6} & -- & -- & -- & -- & -- & \underline{11.6} & -- \\
    \midrule
    \midrule
    \rowcolor{backcolor}
    \multicolumn{13}{c}{\textit{Pre-execution Optimization}} \\
    \midrule
    AutoAgent & 36.0 & 12.0 & 32.0 & 31.5 & -- & -- & 28.0 & 9.0 & 25.0 & \underline{28.5} & -- & -- \\
    \midrule
    \midrule
    \rowcolor{backcolor}
    \multicolumn{13}{c}{\textit{Planner-Worker Orchestration}} \\
    \midrule
    OWL & 37.0 & 12.0 & \underline{36.9} & 29.2 & -- & -- & 23.0 & 7.0 & 23.3 & 21.8 & -- & -- \\
    \midrule
    \midrule
    \rowcolor{backcolor}
    \multicolumn{13}{c}{\textit{Configuration Patching}} \\
    \midrule
    ReSum & 46.5 & 11.0 & 35.9 & -- & -- & -- & \underline{39.5} & 7.0 & 27.2 & -- & -- & -- \\
    OAgents & \underline{48.0} & 12.0 & 30.1 & -- & -- & -- & 33.0 & 9.0 & 22.3 & -- & -- & -- \\
    AgenTRIM & 41.0 & 13.0 & 33.0 & 29.7 & -- & -- & 27.0 & 5.0 & 17.5 & 19.3 & -- & -- \\
    \midrule
    \midrule
    \rowcolor{backcolor}
    \multicolumn{13}{c}{\textit{Ours}} \\
    \midrule
    \name{} & \textbf{61.5} & \textbf{19.0} & \textbf{47.6} & \textbf{43.0} & \textbf{17.8} & \textbf{37.8} & \textbf{49.0} & \textbf{15.0} & \textbf{41.8} & \textbf{35.2} & \textbf{14.1} & \textbf{31.0} \\
    \bottomrule
    \end{tabular}%
    }
    
    \end{table}

\subsection{Experimental Setup}

\textbf{Benchmarks} (Sec.~\ref{sec:benchmarks}).
We evaluate \name{} on five evaluation datasets derived from four long-horizon agent benchmarks, covering diverse task domains including deep research, general AI assistance, and software engineering: FRAMES~\citep{krishna2024fact}, xbench~\citep{chen2025xbench}, GAIA~\citep{mialon2023gaia}, and SWE-bench Lite~\citep{jimenez2023swe}. Following WebSailor~\citep{li2025websailor}, we additionally report GAIA(WS), a subset of 103 validation samples solvable using only search and web browsing. Dataset splits, evaluation metrics, and the MathVista~\citep{lu2023mathvista} data used only as a training supplement are detailed in Appendix~\ref{sec:benchmarks}.

\textbf{Baselines} (Sec.~\ref{sec:baselines}).
We organize the reported baselines according to the taxonomy in the related work, covering general-purpose and task-specialized agents with static configuration, planner-worker orchestration methods such as OWL~\citep{hu2025owl}, and configuration-patching methods such as ReSum~\citep{wu2025resum} and OAgents~\citep{zhu2025oagents}. Baseline definitions, applicable datasets, and fair evaluation settings are provided in Appendix~\ref{sec:baselines}.

\textbf{Implementation} (Sec.~\ref{sec:exp_details}).
We use \texttt{Qwen3-8B} and \texttt{Qwen3-14B} as the base models for the execution agent. CAT consists of RFT-based cold-start initialization followed by KTO-based reinforcement learning refinement, with teacher trajectories generated by \texttt{Qwen3-235B-A22B-Thinking}. Other details are described in Appendix~\ref{sec:exp_details}.

\subsection{Main Results}
\label{sec:main_results}

\textbf{Runtime Specialization enables \name{} to dynamically update its configuration across task stages, outperforming static baselines and pre-execution optimization methods while preserving cross-task generalization.}
Table~\ref{tab:main_results} reports the main results across two model scales and benchmarks spanning deep research, general AI assistance, and software engineering.
Across \texttt{Qwen3-14B} and \texttt{Qwen3-8B}, the static Vanilla Agent averages only 25.9\% and 15.2\%, whereas zero-shot \name{} reaches 37.8\% and 31.0\%, showing that dynamic configuration improves long-horizon tasks broadly rather than on a single benchmark. \name{} also surpasses task-specialized static systems: with \texttt{Qwen3-14B}, it scores 61.5\% on FRAMES versus 42.0\% for WebSailor and 46.0\% for Co-Sight, and 17.8\% on SWE-Lite versus 14.2\% for SWE-Agent and 14.6\% for SWE-Search. Compared with AutoAgent, whose configuration is fixed before execution, \texttt{Qwen3-14B}/\texttt{Qwen3-8B} \name{} achieves 43.0\% and 35.2\% on GAIA, above 31.5\% and 28.5\%. These results show that the main gains come from continuous runtime specialization rather than a one-shot pre-execution choice.

\textbf{Integrated Adaptation and Execution unifies task execution and configuration adaptation within a single policy, yielding stronger gains than decoupled methods.}
Existing decoupled approaches separate adaptation from execution: AutoAgent performs only pre-execution optimization, OWL relies on planner-worker orchestration, and ReSum, OAgents, and AgenTRIM depend on external patches. \name{} consistently outperforms them on long-horizon tasks. With \texttt{Qwen3-14B}, it scores 43.0\% on GAIA, above AutoAgent's 31.5\%, OWL's 29.2\%, and AgenTRIM's 29.7\%; on FRAMES, it reaches 61.5\%, ahead of AutoAgent's 36.0\%, OWL's 37.0\%, ReSum's 46.5\%, and OAgents' 48.0\%. With \texttt{Qwen3-8B}, \name{} also scores 35.2\% on GAIA, above AutoAgent's 28.5\%, OWL's 21.8\%, and AgenTRIM's 19.3\%. This suggests that internalizing configuration adaptation into the execution policy reduces the information loss caused by cross-module relay, scheduling, and patching.

\textbf{Standardized and Comprehensive Updating jointly updates the major configuration dimensions through a unified tool interface, producing more stable gains than fragmented patches to isolated dimensions.}
Patch-based methods remain uneven because they target only part of the configuration: ReSum mainly improves context management, OAgents focuses on planning, and AgenTRIM brings only limited gains on GAIA. In contrast, \name{} jointly updates the sub-goal, execution strategy, toolbox, knowledge, and context-management mode, and remains strong across benchmarks. With \texttt{Qwen3-8B}, \name{} scores 49.0\% on FRAMES, 41.8\% on GAIA(WS), and 15.0\% on xbench, consistently outperforming ReSum, OAgents, and AgenTRIM. This cross-task consistency indicates that the gains come from coordinated multi-dimensional updating under a unified interface, not from isolated patches.

\subsection{Configuration-Aware Two-stage Training}
\label{sec:training_results}

\begin{table}[t]
\caption{Effect of Configuration-Aware Two-stage Training (CAT) on \texttt{Qwen3-8B}. Results show progressive gains from untrained \name{} to RFT cold-start initialization and KTO reinforcement learning refinement. \name{}+RFT applies rejection sampling fine-tuning, and \name{}+RFT+KTO further optimizes the same execution policy with trajectory-level binary feedback. Best Baseline denotes the strongest non-\name{} baseline on each benchmark from Table~\ref{tab:main_results}. Avg.\ denotes the average over the four reported benchmarks. Bold indicates the strongest result and underline indicates the second strongest result in each column.}
\label{tab:training_effect}
\centering
\begingroup
\setlength{\tabcolsep}{4pt}
\begin{tabular}{@{}lccccc@{}}
\toprule
\multirow{2}{*}{\textbf{Method}} & \multicolumn{5}{c}{\texttt{Qwen3-8B}} \\
\cmidrule(lr){2-6}
& FRAMES & xbench & GAIA(WS) & GAIA & Avg. \\
\midrule
\midrule
\rowcolor{backcolor}
\multicolumn{6}{c}{\textit{Reference Baselines}} \\
\midrule
Vanilla Agent & 21.0 & 6.0 & 19.4 & 19.7 & 16.5 \\
Best Baseline & 39.5 & 13.0 & 28.1 & 28.5 & 27.3 \\
\midrule
\midrule
\rowcolor{backcolor}
\multicolumn{6}{c}{\textit{Ours}} \\
\midrule
\name{} & 49.0 & 15.0 & 41.8 & 35.2 & 35.3 \\
\name{} + RFT & \underline{54.0} & \underline{23.0} & \underline{43.7} & \underline{35.8} & \underline{39.1} \\
\name{} + RFT + KTO & \textbf{58.5} & \textbf{28.0} & \textbf{50.5} & \textbf{44.2} & \textbf{45.3} \\
\bottomrule
\end{tabular}
\endgroup

\end{table}

\textbf{By unifying configuration updates and task execution within the same policy, \name{} supports Joint Optimization; after CAT training, \name{} raises average accuracy from 16.5\% for the static-configuration baseline to 45.3\%, a 28.8-point absolute gain.}
Table~\ref{tab:training_effect} compares untrained \name{}, RFT, and KTO-based reinforcement learning on \texttt{Qwen3-8B} against Vanilla Agent and the strongest non-\name{} baselines. Untrained \name{} already averages 35.3\%, substantially above Vanilla Agent's 16.5\% and the 27.3\% average of the strongest non-\name{} baselines. CAT further lifts the average to 45.3\%, with especially large gains on xbench, from 15.0\% to 28.0\%, and GAIA, from 35.2\% to 44.2\%, indicating stronger complex search, cross-stage reasoning, and configuration updating. With RFT and KTO, \name{} achieves the best results across all benchmarks, including 58.5\% on FRAMES. This confirms that CAT jointly optimizes execution and reconfiguration rather than isolated trajectories.

\textbf{CAT internalizes runtime self-reconfiguration from an invocable mechanism into a policy-level capability, enabling the emergence of adaptive capabilities, including long-horizon planning, self-refinement, and cross-stage strategy optimization.}
On \texttt{Qwen3-8B}, \name{} improves from 35.3\% to 39.1\% after RFT and to 45.3\% after KTO, with GAIA reaching 44.2\%, indicating that KTO substantially improves cross-stage configuration updating.
For trajectory analysis, we concatenate stage-level sub-goals and evaluate them with LLM-as-Judge; the evaluation settings and prompt are provided in Appendix~\ref{sec:trajectory_capability_judge_prompt}, and the full results are reported in Appendix Table~\ref{tab:adaptive_capabilities}.
We find that long-horizon planning is already stable at around 80\% before training.
The main qualitative shift under CAT is the emergence of closed-loop feedback: intermediate verification rises from 70.3\% to 84.8\%, and adaptive strategy adjustment from 35.2\% to 51.5\%.
The GAIA case studies in Appendix~\ref{sec:case_studies}, both drawn from \texttt{Qwen3-8B} \name{}+RFT+KTO trajectories, further show the specific behaviors internalized by CAT: stage-wise long-horizon decomposition, feedback-driven strategy pivot after failed evidence extraction, cross-stage knowledge accumulation, verification before output, and self-determined termination.
CAT therefore improves not only final accuracy, but also replaces manually injected execution guidance with emergent adaptivity that verifies progress and adjusts strategy during execution.

\subsection{Ablation Study}
\label{sec:ablation}

\textbf{Integrated configuration generation consistently outperforms the decoupled variant across both model scales and all benchmarks, demonstrating that the unified treatment of task execution and configuration generation is a key driver of \name{}'s gains.}
We introduce a decoupled variant of \name{}, denoted as Decoupled, which delegates next-stage configuration generation to a separate LLM with the same backbone, using only the configuration-generation specification without access to the execution agent's full context. The original \name{} design is denoted as Integrated. Appendix Table~\ref{tab:decouple_ablation} quantifies this advantage: Integrated surpasses Decoupled by 4.3 and 6.1 points on \texttt{Qwen3-14B} and \texttt{Qwen3-8B}, respectively. It improves all five evaluation datasets, most notably in long-horizon, multi-capability tasks such as the GAIA series. This suggests that next-stage configuration generation depends heavily on fine-grained execution states; once delegated to an external LLM operating only on the specification, important information about task progress, knowledge gaps, tool bottlenecks, and context state is lost through relay and compression. By unifying adaptation and execution within the same policy, \name{} reduces coordination loss and improves the quality of runtime specialization.

\textbf{Further analysis attributes \name{}'s gains to a learnable self-reconfiguration mechanism within a unified action space, rather than to extra compute, incidental alignment to a particular model family, or run-to-run randomness.}
Specifically, Sec.~\ref{sec:dynamic_vs_static_analysis} compares \name{} with a Vanilla Agent that lacks the reconfiguration tool and validates the advantage of the unified action space in both accuracy and token efficiency. Sec.~\ref{sec:component_ablation_analysis} decomposes the major configuration dimensions and identifies knowledge transfer and sub-goal planning as the most critical factors for long-horizon reasoning. Sec.~\ref{sec:training_analysis_appendix} separates the effects of RFT and KTO, attributing the gains to improved configuration decisions rather than mere exploration. Sec.~\ref{sec:reconfig_budget_analysis} relates reconfiguration budget to performance and motivates larger adaptive budgets for more complex tasks. Sec.~\ref{sec:tool_selection_analysis} analyzes task-adaptive tool selection on GAIA, revealing that within-stage dynamic adaptation to local task demands also emerges in \name{}. Sec.~\ref{sec:minimax_generalization_analysis} establishes that the mechanism generalizes across model families, including \texttt{MiniMax-M2.5}, while Sec.~\ref{sec:run_variance_analysis} documents that the observed gains remain stable across independent runs.

\section{Conclusion}
\label{sec:Conclusion}

We propose \name{}, a tool-driven paradigm that unifies task execution and runtime self-reconfiguration within a single optimizable policy. By turning configuration updates into native actions, \name{} replaces manually injected guidance with emergent adaptability learned from trajectory-level feedback. With CAT training, \name{} gains 28.8 points over the static-configuration baseline on average and shows that self-adaptive agents can rival specialized workflows while preserving generalist flexibility.

\clearpage
\bibliography{example_paper}
\bibliographystyle{plainnat}

\newpage
\appendix
\section{Appendix}
\label{sec:Appendix}

\subsection{Impact Statement}
\label{sec:impact_statement}

This work advances LLM-based agentic systems by enabling runtime self-reconfiguration for long-horizon tasks. Potential positive impacts include improving the robustness and generality of agents for research assistance, software engineering, and complex information-seeking workflows. Potential negative impacts include more capable autonomous tool use, web search, and code modification, which may amplify errors, unsafe actions, or misuse if deployed without oversight. This work is a research contribution rather than a deployed system. Practical deployment should include human supervision, task-level constraints, sandboxed tool execution, and monitoring for unsafe or unintended behavior.

\subsection{Related Work}
\label{sec:related_work}

As discussed in Sec.~\ref{sec:Introduction}, the \emph{configuration} of an agentic system defines how the agent reasons, acts, and interacts with its environment during execution, typically including key elements such as execution strategy, toolbox, context, and context-management mode. Existing work has extensively studied configuration design for long-horizon tasks, yet its central challenge lies in the tension between domain-specific performance and generality: high-performing systems often depend on task-specialized configurations, whereas general-purpose systems must rely on task-agnostic workflows, broad toolsets, and generic context strategies to cover diverse tasks. Prior methods mainly address this tension through pre-execution optimization, planner-worker orchestration, and configuration patching. However, these approaches generally decouple task adaptation from task execution, limiting the ability of configuration updates to capture execution details and making unified credit assignment with execution behavior difficult.

\textbf{Agents with static configuration: general-purpose and task-specialized agents.}
General-purpose single agents, represented by ReAct~\citep{yao2022react}, combine reasoning and tool invocation through a thought-action-observation loop, thereby supporting open-ended task execution.
OpenHands~\citep{wang2025openhands} further provides composable and extensible infrastructure for general-purpose agents, improving their deployability in real development environments.
Despite their generality, these systems typically rely on configurations fixed before execution, including predefined reasoning formats, tool interfaces, and context-management strategies.

In contrast, many high-performing agents or workflows improve task performance by encoding stronger domain priors into their configurations.
SWE-Agent~\citep{yang2024swe} designs an Agent-Computer Interface for software engineering, while SWE-Search~\citep{antoniades2024swe} uses Monte Carlo tree search to explore the solution space of software engineering tasks.
WebSailor~\citep{li2025websailor}, Search-o1~\citep{li2025search}, and Co-Sight~\citep{zhang2025cosight} design specialized search, browsing, and reasoning procedures for deep research or search-enhanced reasoning.
Together, these two lines of static configuration design expose a fundamental trade-off: stronger domain priors improve task-specific performance but couple the system more tightly to a particular task distribution, whereas more general configurations improve transferability at the cost of broader action spaces, weaker task guidance, and reduced efficiency and robustness in long-horizon tasks~\citep{xia2024agentless}.

\textbf{Pre-execution optimization and automated agent construction.}
Another line of work seeks to automatically obtain better agent or workflow configurations before execution. AFlow~\citep{zhang2024aflow} automatically generates agentic workflows, AgentSquare~\citep{shang2024agentsquare} searches for suitable agent structures within a modular design space, and AutoAgent~\citep{tang2025autoagent} proposes a no-code framework for creating and deploying LLM agents through natural language. These methods reduce the cost of manually designing configurations and can yield stronger task-specialized structures prior to execution. However, their optimized outputs are typically instantiated before runtime as relatively fixed agents, workflows, or configuration templates. Once execution begins, the execution agent itself usually cannot directly convert environmental feedback, stage progress, or failure signals into intrinsic configuration updates. Consequently, they remain limited in addressing the evolving sub-goals, tool requirements, and context-management needs of long-horizon tasks.

\textbf{Planner-worker orchestration.}
Planner-worker orchestration employs a centralized planner or orchestrator to coordinate multiple specialized workers, expanding task coverage while preserving domain capabilities. OWL~\citep{hu2025owl} adopts a hierarchical architecture in which a Planner decomposes tasks, a Coordinator dispatches them to workers such as Web Agent and Coding Agent, and replanning is performed after failures. Magentic-One~\citep{fourney2024magentic} follows a similar orchestrator-workers pattern. Although such orchestration can cover heterogeneous task types through division of labor, task adaptation is mainly handled by the planner or orchestrator, while actual execution is delegated to workers. Task states must therefore be repeatedly relayed, compressed, and summarized among planners, coordinators, and workers, making information loss and error accumulation likely. Moreover, final success or failure is difficult to attribute cleanly to the planner's adaptation decisions versus the workers' execution behaviors, hindering end-to-end joint optimization.

\textbf{Configuration patching.}
Recent single-agent extensions patch individual dimensions of a general agent's configuration through external mechanisms. ReSum~\citep{wu2025resum} mitigates context overload in long-horizon search through periodic context summarization, while MEM1~\citep{zhou2025mem1} learns coordinated memory and reasoning strategies to improve long-horizon efficiency. OAgents~\citep{zhu2025oagents} uses an external LLM planner to periodically update the agent's plan during execution, and AdaPlanner~\citep{sun2023adaplanner} performs adaptive planning based on environmental feedback. Along the tool dimension, AgenTRIM~\citep{betser2026agentrim} dynamically trims, filters, and validates the tool set exposed to the agent, reducing irrelevant or unsuitable tool use during execution. These methods provide useful patches for individual dimensions such as context, plan, or toolbox, but they typically depend on external modules or fixed triggering mechanisms and only modify local aspects of the configuration. Because task adaptation is not unified into the execution agent's native action space, such methods struggle to holistically coordinate the major configuration dimensions and are difficult to optimize end to end through a unified trajectory.

\textbf{Reinforcement learning and emergent reasoning behavior.}
Recent work such as DeepSeek-R1~\citep{guo2025deepseek} shows that reinforcement learning can incentivize advanced reasoning behaviors in LLMs.
This observation is closely related to our training motivation, but \name{} applies reinforcement learning to a different object: rather than only optimizing token-level reasoning traces, CAT optimizes trajectory-level configuration decisions that govern how the agent plans sub-goals, selects strategies and tools, manages context, and continues execution across stages.
Thus, in \name{}, RL is not used to reinforce an externally handcrafted mechanism, but to optimize the agent's native capability for runtime configuration adaptation.

In contrast to prior work, \name{} abstracts configuration updates as a standardized tool interface and unifies task adaptation and task execution within the single action space of one policy. The execution agent no longer passively receives configuration changes from an external optimizer, planner, or patching module; instead, it actively invokes the reconfiguration tool based on task progression and environmental feedback, comprehensively updating the sub-goal, strategy, toolbox, context, and context-management mode. This places configuration decisions and execution behaviors on the same trajectory, reducing information loss from cross-module relay and compression while allowing downstream success or failure to serve directly as trajectory-level credit assignment. As a result, \name{} achieves runtime self-adaptation and end-to-end joint optimization for long-horizon tasks while preserving generalist flexibility.

\subsection{Additional Experimental Analysis}
\label{sec:additional_analysis}

\begin{table}[H]
\caption{Emergent adaptive capabilities under CAT on GAIA using \texttt{Qwen3-8B}. We concatenate stage-level sub-goals into sub-goal trajectories and use LLM-as-Judge to evaluate long-horizon task planning, self-refinement, self-correction, verification, and strategy adaptation.}
\label{tab:adaptive_capabilities}
\centering
\footnotesize
\setlength{\tabcolsep}{3pt}
\begin{tabular}{@{}lccccc@{}}
\toprule
\textbf{Run} & \textbf{\shortstack[c]{Long-horizon\\Task Planning}} & \textbf{\shortstack[c]{Self-\\refinement}} & \textbf{\shortstack[c]{Self-\\correction}} & \textbf{Verification} & \textbf{\shortstack[c]{Strategy\\Adaptation}} \\
\midrule
\name{} & 80.0\% & 55.2\% & 40.0\% & 70.3\% & 35.2\% \\
\name{} + RFT & 81.2\% & 53.9\% & 43.0\% & 80.0\% & 38.2\% \\
\name{} + RFT + KTO & 80.6\% & 64.2\% & 49.1\% & 84.8\% & 51.5\% \\
\bottomrule
\end{tabular}
\end{table}

\subsubsection{Decoupled vs. Integrated Configuration Generation}
\label{sec:decouple_ablation_appendix}

\begin{table}[H]
\caption{Ablation study on integrated configuration generation. The decoupled variant delegates next-stage configuration generation to a separate LLM with the same backbone, using only the configuration-generation specification but without the execution agent's full context. Integrated denotes the default \name{} design, where configuration generation is performed within the execution agent's context.}
\label{tab:decouple_ablation}
\centering
\footnotesize
\begin{tabular}{l l c c c c c c}
\toprule
\textbf{Model} & \textbf{Version} & \textbf{FRAMES} & \textbf{xbench} & \textbf{GAIA(WS)} & \textbf{GAIA} & \textbf{SWE-Lite} & \textbf{Avg.} \\
\midrule
\texttt{Qwen3-14B} & Decoupled & 56.0 & 16.0 & 40.8 & 38.8 & 16.1 & 33.5 \\
\texttt{Qwen3-14B} & Integrated & 61.5 & 19.0 & 47.6 & 43.0 & 17.8 & 37.8 \\
\texttt{Qwen3-8B} & Decoupled & 44.0 & 10.0 & 30.1 & 27.9 & 12.4 & 24.9 \\
\texttt{Qwen3-8B} & Integrated & 49.0 & 15.0 & 41.8 & 35.2 & 14.1 & 31.0 \\
\bottomrule
\end{tabular}
\end{table}

\subsubsection{Unified Action Space vs. Static Execution}
\label{sec:dynamic_vs_static_analysis}

\begin{table}[H]
\caption{Unified action space paradigm vs. passive execution paradigm on GAIA using \texttt{Qwen3-8B}. Comparison of \name{} and Vanilla Agent across accuracy (\%), average execution steps, maximum input tokens, and total token consumption. Vanilla Agent denotes a standard ReAct-style agent equivalent to the inference policy without the reconfiguration tool.}
\label{tab:dynamic_vs_static}
\centering
\footnotesize
\setlength{\tabcolsep}{5pt}
\begin{tabular}{l l c c c c}
\toprule
\textbf{Method} & \textbf{Level} & \textbf{Acc.} & \textbf{Avg. Steps} & \textbf{Max Input Tokens} & \textbf{Total Tokens} \\
\midrule
Vanilla Agent & All & 19.7 & 7.96 & 9527.5 & 9527.5 \\
Vanilla Agent & Level 1 & 22.6 & 7.85 & 8866.1 & 8866.1 \\
Vanilla Agent & Level 2 & 23.3 & 8.10 & 10288.2 & 10288.2 \\
Vanilla Agent & Level 3 & 3.85 & 7.56 & 8999.6 & 8999.6 \\
\name{} & All & 35.2 & 13.27 & 14656.94 & 30315.2 \\
\name{} & Level 1 & 45.3 & 5.74 & 13289.05 & 11550.0 \\
\name{} & Level 2 & 36.0 & 15.44 & 15489.93 & 39384.7 \\
\name{} & Level 3 & 11.5 & 33.63 & 15219.38 & 70991.8 \\
\bottomrule
\end{tabular}
\end{table}

\textbf{Unified-action-space adaptation substantially outperforms the static configuration paradigm in both effectiveness and efficiency.} Table~\ref{tab:dynamic_vs_static} compares \name{} and Vanilla Agent on GAIA with \texttt{Qwen3-8B} across difficulty levels, reporting accuracy, average execution steps, maximum input length, and total token consumption. \name{} improves overall accuracy from 19.7\% to 35.2\% over Vanilla Agent, identifying self-reconfiguration as a central source of the gain. The model allocates more exploration budget to harder tasks, with the largest increase in average steps occurring at Level 3, while keeping the maximum input length within a controlled range. Notably, on the easier Level 1 subset, \name{} achieves higher accuracy with fewer average steps, suggesting that self-reconfiguration also improves execution efficiency on simpler tasks. Although total token usage increases with deeper exploration, the maximum input length remains bounded at roughly 13K to 15K tokens, reflecting effective history control through agent-driven context management. By comparison, Co-Sight consumes about 229K/261K/315K tokens per task on GAIA Level 1/2/3, whereas \name{} uses only 11.6K/39.4K/71.0K, corresponding to roughly 19.8/6.6/4.4 times higher token efficiency. Together, these results connect the unified action space to gains in both accuracy and computational efficiency on long-horizon tasks.

\subsubsection{Component Ablation}
\label{sec:component_ablation_analysis}

\begin{table}[H]
\caption{Ablation study on configuration components using \texttt{Qwen3-8B}. Impact of removing individual reconfiguration elements on GAIA performance (\%). \texttt{w/o} denotes \texttt{without}.}
\label{tab:component_ablation}
\centering
\footnotesize
\setlength{\tabcolsep}{5pt}
\begin{tabular}{l c c c c c c}
\toprule
\textbf{Level} & \textbf{Full} & \textbf{w/o Sub-goal} & \textbf{w/o Strategy} & \textbf{w/o Toolbox} & \textbf{w/o Knowledge} & \textbf{w/o Context Mgmt.} \\
\midrule
All & 35.2 & 30.9 & 34.5 & 33.3 & 27.3 & 33.9 \\
Level 1 & 45.3 & 43.4 & 47.2 & 41.5 & 47.2 & 45.3 \\
Level 2 & 36.0 & 31.4 & 34.9 & 37.2 & 22.1 & 34.9 \\
Level 3 & 11.5 & 3.85 & 7.69 & 3.85 & 3.85 & 7.69 \\
\bottomrule
\end{tabular}
\end{table}

\textbf{The configuration components contribute unevenly to performance, with knowledge transfer and sub-goal planning emerging as the most critical.} Table~\ref{tab:component_ablation} presents a component-wise ablation on \texttt{Qwen3-8B}, showing how removing sub-goal planning, strategy, toolbox, knowledge, or context management affects GAIA accuracy across difficulty levels. Removing knowledge is the most damaging intervention, reducing overall accuracy from 35.2\% to 27.3\% and Level 2 performance from 36.0\% to 22.1\%. This highlights the importance of cross-stage information accumulation for long-horizon reasoning. Removing sub-goal planning also substantially degrades performance, lowering overall accuracy to 30.9\% and Level 3 accuracy from 11.5\% to 3.85\%, which underscores the importance of macro-level planning. Strategy, toolbox, and context management have smaller effects on the aggregate score, but all weaken Level 3 performance, pointing to complementary roles in harder tasks.

\subsubsection{Training-Stage Analysis}
\label{sec:training_analysis_appendix}

\begin{table}[H]
\caption{Detailed analysis of different training stages on GAIA using \texttt{Qwen3-8B}. Accuracy (\%), average execution steps, and reconfiguration iterations across difficulty levels. \name{}+RFT applies rejection sampling fine-tuning for cold-start initialization; \name{}+RFT+KTO further incorporates KTO reinforcement learning.}
\label{tab:training_analysis}
\centering
\footnotesize
\setlength{\tabcolsep}{4pt}
\resizebox{\linewidth}{!}{%
\begin{tabular}{l c c c c c c c c c}
\toprule
\textbf{Level} & \textbf{\name{} Acc.} & \textbf{\name{} Steps} & \textbf{\name{} Iters} & \textbf{+RFT Acc.} & \textbf{+RFT Steps} & \textbf{+RFT Iters} & \textbf{+RFT+KTO Acc.} & \textbf{+RFT+KTO Steps} & \textbf{+RFT+KTO Iters} \\
\midrule
All & 35.2 & 13.27 & 6.33 & 35.8 & 13.43 & 6.64 & 44.2 & 11.65 & 5.66 \\
Level 1 & 45.3 & 5.74 & 4.58 & 56.6 & 4.01 & 4.04 & 56.6 & 4.37 & 3.71 \\
Level 2 & 36.0 & 15.44 & 6.95 & 31.4 & 19.32 & 7.97 & 44.2 & 12.60 & 6.29 \\
Level 3 & 11.5 & 33.63 & 10.25 & 7.69 & 32.30 & 9.27 & 19.2 & 42.92 & 10.07 \\
\bottomrule
\end{tabular}
}
\end{table}

\textbf{Joint training improves both task performance and configuration efficiency.} Table~\ref{tab:training_analysis} breaks down the performance of \name{}, \name{}+RFT, and \name{}+RFT+KTO on GAIA, reporting accuracy, average execution steps, and reconfiguration iterations across difficulty levels. \texttt{+RFT} alone yields only limited overall gains and even degrades performance on harder subsets, whereas \texttt{+RFT+KTO} raises overall accuracy from 35.2\% to 44.2\% while reducing average steps from 13.27 to 11.65 and reconfiguration iterations from 6.33 to 5.66. The gains are larger on harder tasks, with Level 2 improving from 36.0\% to 44.2\% and Level 3 from 11.5\% to 19.2\%. This suggests that reinforcement learning primarily improves the quality of long-horizon configuration decisions rather than merely increasing exploration.

\subsubsection{Reconfiguration-Budget Sensitivity}
\label{sec:reconfig_budget_analysis}

\begin{table}[H]
\caption{Performance of \texttt{Qwen3-8B} \name{} under different maximum reconfiguration caps on GAIA.}
\label{tab:reconfig_budget}
\centering
\footnotesize
\begin{tabular}{l c c c c c}
\toprule
\textbf{Level} & \textbf{Final Acc.} & \textbf{update$\leq$2} & \textbf{update$\leq$8} & \textbf{update$\leq$16} & \textbf{update$\leq$30} \\
\midrule
Total & 35.15 & 18.18 & 32.73 & 33.94 & 35.15 \\
Level 1 & 45.28 & 26.42 & 45.28 & 45.28 & 45.28 \\
Level 2 & 36.05 & 17.44 & 32.56 & 34.88 & 36.05 \\
Level 3 & 11.54 & 3.85 & 7.69 & 7.69 & 11.54 \\
\bottomrule
\end{tabular}
\end{table}

\textbf{Performance improves monotonically as the maximum number of reconfigurations increases.} Table~\ref{tab:reconfig_budget} reports how the final accuracy of \texttt{Qwen3-8B} \name{} changes under different caps on the maximum number of reconfigurations. Overall accuracy rises from 18.18\% at \texttt{update<=2} to 35.15\% at \texttt{update<=30}. The saturation pattern differs by difficulty: Level 1 reaches its final performance by 8 updates, whereas Level 2 continues improving through 30 updates and Level 3 still benefits from the largest budget. This indicates that reconfiguration is not redundant overhead, but an effective adaptive resource whose value grows with task complexity.

\subsubsection{Task-Adaptive Tool Selection}
\label{sec:tool_selection_analysis}

\begin{table}[H]
\caption{Task-adaptive tool selection patterns on GAIA using \texttt{Qwen3-8B}. Tool selection probabilities (\%) across different task types demonstrate \name{}'s ability to dynamically configure task-appropriate toolboxes through self-reconfiguration. \texttt{code interp.} abbreviates \texttt{code\_interpreter}.}
\label{tab:tool_selection_patterns}
\centering
\footnotesize
\setlength{\tabcolsep}{5pt}
\begin{tabular}{l c c c c c c}
\toprule
\textbf{Task Type} & \textbf{\texttt{visit}} & \textbf{\texttt{search}} & \textbf{\shortstack[c]{\texttt{code}\\\texttt{interp.}}} & \textbf{\texttt{bash}} & \textbf{\shortstack[c]{\texttt{str}\\\texttt{replace}}} & \textbf{\shortstack[c]{\texttt{file}\\\texttt{analyzer}}} \\
\midrule
Web browsing & 80.1 & 82.7 & 48.4 & 11.4 & 18.7 & 65.0 \\
Coding & 37.5 & 36.7 & 94.4 & 8.9 & 74.9 & 59.9 \\
Multi-modality & 8.8 & 10.8 & 29.5 & 94.5 & 67.4 & 84.0 \\
Diverse filetype & 23.1 & 21.5 & 37.9 & 61.3 & 77.3 & 94.5 \\
\bottomrule
\end{tabular}
\end{table}

\textbf{\name{}'s adaptivity is not limited to cross-stage configuration updates; within individual stages, it also exhibits dynamic adaptation to the current task demands.} Table~\ref{tab:tool_selection_patterns} reports the tool selection probabilities of \texttt{Qwen3-8B} \name{} on GAIA across four task types: web browsing, coding, multi-modality, and diverse filetype tasks. The resulting tool usage is not a fixed template, but changes substantially with task type. Web browsing tasks most frequently select \texttt{visit} and \texttt{search}, coding tasks strongly favor \texttt{code\_interpreter} and \texttt{str\_replace\_editor}, while multi-modality and diverse filetype tasks rely more heavily on \texttt{file\_analyzer}, \texttt{execute\_bash}, and file-processing tools. These differences indicate that \name{} not only updates long-horizon configurations across stages, but also selects more appropriate toolboxes according to the local task state within each stage. If the agent merely used a static tool set, tool distributions across task types would be expected to be more similar; instead, the table shows clear task-dependent tool preferences, suggesting that self-reconfiguration has been internalized as fine-grained dynamic task adaptation during execution.

\subsubsection{Cross-Family Generalization}
\label{sec:minimax_generalization_analysis}

\begin{table}[H]
\caption{GAIA comparison between Vanilla Agent and \name{} on \texttt{MiniMax-M2.5} (230B MoE, 10B activated).}
\label{tab:minimax_generalization}
\centering
\footnotesize
\begin{tabular}{l c c c c}
\toprule
\textbf{Model} & \textbf{Overall} & \textbf{Level 1} & \textbf{Level 2} & \textbf{Level 3} \\
\midrule
Vanilla Agent & 33.3 & 41.5 & 30.2 & 15.4 \\
\name{} & 44.2 & 56.6 & 41.9 & 26.9 \\
\bottomrule
\end{tabular}
\end{table}

\textbf{On \texttt{MiniMax-M2.5}, \name{} consistently outperforms Vanilla Agent across all difficulty levels.} Table~\ref{tab:minimax_generalization} compares \name{} and Vanilla Agent on \texttt{MiniMax-M2.5} in terms of GAIA accuracy, both overall and by difficulty level. \name{} improves overall GAIA accuracy from 33.3\% to 44.2\%. The gains are not confined to a single subset, but extend across Level 1, Level 2, and Level 3 simultaneously. This supports the cross-family generality of self-reconfiguration beyond the Qwen series, extending it to a large-scale MoE model with substantially different architecture and scale.

\subsubsection{Run-to-Run Stability}
\label{sec:run_variance_analysis}

\begin{table}[H]
\caption{Mean, standard deviation, and coefficient of variation over five independent runs of off-the-shelf \texttt{Qwen3-14B} \name{} across benchmarks.}
\label{tab:run_variance}
\centering
\footnotesize
\begin{tabular}{l c c}
\toprule
\textbf{Benchmark} & \textbf{Mean $\pm$ Std} & \textbf{CV (\%)} \\
\midrule
FRAMES & $59.8 \pm 3.01$ & 5.0 \\
xbench & $17.8 \pm 2.28$ & 12.8 \\
GAIA(WS) & $45.6 \pm 4.61$ & 10.1 \\
GAIA & $41.8 \pm 3.88$ & 9.3 \\
SWE-bench Lite & $17.3 \pm 2.01$ & 11.6 \\
\bottomrule
\end{tabular}
\end{table}

\textbf{Variance over five independent runs remains within an acceptable range.} Table~\ref{tab:run_variance} reports the mean, standard deviation, and coefficient of variation over five independent runs of off-the-shelf \texttt{Qwen3-14B} \name{} across benchmarks. The coefficients of variation fall between 5.0\% and 12.8\%. FRAMES is the most stable benchmark, while xbench and SWE-bench Lite exhibit relatively larger but still reasonable variation. Overall, the gains of \name{} remain consistent across runs, pointing to mechanism-level improvements rather than a few favorable samples.

\subsection{Case Studies}
\label{sec:case_studies}

To illustrate how \name{} behaves under different settings, we present two CAT-trained GAIA trajectories and one zero-shot SWE-bench Lite trajectory. The GAIA cases highlight the policy-level behaviors internalized after CAT, while the SWE case shows how the same reconfiguration mechanism supports software debugging and repair without task-specific training.

\subsubsection{CAT-trained GAIA Trajectories}

Table~\ref{tab:case_evolution} summarizes two representative GAIA trajectories generated by \texttt{Qwen3-8B} \name{}+RFT+KTO. The \emph{Knowledge Update} column records the new task facts produced in each stage and carried forward through later configurations.

\begin{table}[H]
\caption{Configuration evolution across stages in two GAIA trajectories generated by \texttt{Qwen3-8B} \name{}+RFT+KTO. The \emph{Knowledge Update} column records new facts produced in each stage and carried forward through $K_i$. ``reconfig'' indicates invoking $T_{\text{reconfig}}$; ``term'' indicates invoking $T_{\text{term}}$.}
\label{tab:case_evolution}
\centering
\resizebox{\linewidth}{!}{%
\begin{tabular}{clllc}
\toprule
\textbf{Stage} & \textbf{Sub-goal Focus} & \textbf{Toolbox} & \textbf{Knowledge Update} & \textbf{Decision} \\
\midrule
\midrule
\rowcolor{backcolor}
\multicolumn{5}{c}{\textit{Case 1: NASA Astronaut Task (Level 3)}} \\
\midrule
1 & Identify astronauts in APOD image & visit, code\_interp., str\_replace, search & Conrad is smaller; Bean also identified & reconfig \\
2 & Determine Conrad's group & visit, code\_interp., \textbf{execute\_bash}, search & Conrad selected in Group 2 & reconfig \\
3 & List Group 2 members & maintained & Nine Group 2 members identified & reconfig \\
4 & Collect space-time data & maintained & Ed White: 5876 min; Elliot See: 0 min & reconfig \\
5 & Verify and output answer & visit, code\_interp., search & \texttt{White;5876} verified & \textbf{term} \\
\midrule
\midrule
\rowcolor{backcolor}
\multicolumn{5}{c}{\textit{Case 2: ASEAN Distance Calculation (Level 2)}} \\
\midrule
1 & Identify ASEAN members and capitals & visit, code\_interp., execute\_bash, search & 10 members found; capitals absent & reconfig \\
2 & Retry with explicit URL/parsing & maintained & ASEAN page parsed; capitals still absent & reconfig \\
3 & \textbf{Pivot}: visit country pages & maintained & All 10 capitals extracted & reconfig \\
4 & Retrieve capital coordinates & maintained & Coordinates collected in decimal degrees & reconfig \\
5 & Compute Haversine distances & \textbf{code\_interp., execute\_bash} & Indonesia--Myanmar verified as farthest & \textbf{term} \\
\bottomrule
\end{tabular}%
}
\end{table}

\paragraph{Case 1: NASA Astronaut Task (GAIA Level 3).}

\textbf{Task Description.} \textit{In NASA's Astronomy Picture of the Day on 2006 January 21, two astronauts are visible, with one appearing much smaller than the other. As of August 2023, out of the astronauts in the NASA Astronaut Group that the smaller astronaut was a member of, which one spent the least time in space, and how many minutes did he spend in space, rounded to the nearest minute? Exclude any astronauts who did not spend any time in space. Give the last name of the astronaut, separated from the number of minutes by a semicolon.}

\textbf{Task Information.} Difficulty: Level 3 (Highest); Total Stages: 5; Final Answer: \texttt{White;5876} $\checkmark$

\textbf{Key Observations.} This CAT-trained trajectory demonstrates a concrete long-horizon planning pattern: image identification $\rightarrow$ group lookup $\rightarrow$ member enumeration $\rightarrow$ data collection $\rightarrow$ verification. It also shows cross-stage knowledge accumulation, as Conrad's identity and Group 2 membership are carried forward until the agent verifies \texttt{White;5876} and autonomously invokes $T_{\text{term}}$.

\paragraph{Case 2: ASEAN Capital Distance Calculation (GAIA Level 2).}

\textbf{Task Description.} \textit{In terms of geographical distance between capital cities, which 2 countries are the furthest from each other within the ASEAN bloc according to wikipedia? Answer using a comma separated list, ordering the countries by alphabetical order.}

\textbf{Task Information.} Difficulty: Level 2; Total Stages: 5; Final Answer: \texttt{Indonesia, Myanmar} $\checkmark$

\textbf{Key Observations.} This CAT-trained trajectory highlights feedback-driven strategy adaptation and closed-loop verification. After the ASEAN page and explicit parsing both fail to provide capitals, the agent pivots to individual country pages, then transitions from retrieval to computation by pruning tools once coordinates are collected, and finally verifies the farthest pair before terminating.

\subsubsection{Zero-shot SWE-bench Lite Trajectory}

As a complementary example, Table~\ref{tab:swe_case} presents a zero-shot \name{} trajectory on a software engineering task.

\textbf{Task Description.} Fix \texttt{ManagementUtility}, which instantiates \texttt{CommandParser} without passing \texttt{prog=self.prog\_name}.

\textbf{Task Information.} Setting: zero-shot \name{}; Benchmark: SWE-bench Lite; Case: \texttt{django\_\_django-13658}; Total Stages: 4; Outcome: patch verified by test $\checkmark$

\begin{table}[H]
\caption{Stage-level sub-goals, toolbox, knowledge updates, and reconfiguration decisions of zero-shot \name{} on the SWE-bench Lite case \texttt{django\_\_django-13658}. \texttt{bash} and \texttt{editor} denote \texttt{execute\_bash} and \texttt{str\_replace\_editor}.}
\label{tab:swe_case}
\centering
\scriptsize
\setlength{\tabcolsep}{3pt}
\begin{tabular}{clllc}
\toprule
\textbf{Stage} & \textbf{Sub-goal} & \textbf{Toolbox} & \textbf{Knowledge Update} & \textbf{Decision} \\
\midrule
1 & Repository setup & \texttt{bash}, \texttt{editor} & Repository root confirmed & reconfig \\
2 & Locate, patch, verify & maintained & File found; patch applied; \texttt{argv[0]=None} crash recovered & reconfig \\
3 & Revert over-fix & maintained & Redundant guard reverted & reconfig \\
4 & Final verification & maintained & Patch verified by test & term \\
\bottomrule
\end{tabular}
\end{table}

\textbf{Key Observations.} This zero-shot SWE example shows adaptation through sub-goal and knowledge updates. The agent moves from environment setup to localization, patching, failed verification, rollback of an over-fix, and final confirmation, illustrating self-diagnosis, self-correction, and self-refinement in code repair.

\subsection{Benchmarks}
\label{sec:benchmarks}

We evaluate \name{} on five evaluation datasets derived from four long-horizon agent benchmarks: FRAMES and xbench for deep research, GAIA and its GAIA(WS) subset for general AI assistant tasks, and SWE-bench Lite for software engineering.

\textbf{FRAMES}~\citep{krishna2024fact} (Factuality, Retrieval, And reasoning MEasurement Set) is a deep research benchmark with 824 questions, designed to assess multi-document retrieval, information integration, and complex reasoning. Each question typically requires retrieving and combining dispersed evidence from 2--15 Wikipedia articles, covering reasoning types such as numerical computation, table parsing, multi-constraint satisfaction, temporal reasoning, and post-processing. We randomly sample 200 questions as the test set and use the remaining 624 as the training pool for constructing CAT's RFT and KTO data.

\textbf{xbench}~\citep{chen2025xbench} evaluates realistic deep search and multi-step retrieval capabilities. We use the DeepSearch-2510 version, which contains 100 deep research questions. The tasks require agents to iteratively locate, filter, and integrate key evidence from large candidate information spaces, thereby testing long-horizon search, tool use, evidence selection, and answer synthesis.

\textbf{GAIA}~\citep{mialon2023gaia} (General AI Assistants Benchmark) evaluates general AI assistant capabilities. We use the validation set with publicly available answers, covering real-world question-answering scenarios in domains such as science, history, and geography across three difficulty levels (Level 1/2/3). Many GAIA questions require task decomposition, web search, calculation, code execution, and multi-source information integration. Following WebSailor~\citep{li2025websailor}, we further curate 103 validation samples solvable using only search and web browsing, denoted as GAIA(WS), for more direct comparison with search-oriented baselines.

\textbf{SWE-bench Lite}~\citep{jimenez2023swe} contains 300 software engineering tasks from real GitHub repositories and evaluates an agent's ability to understand codebases, localize bugs, generate patches, and pass relevant tests. Each instance includes a problem description, a codebase, and test cases. The evaluation metric is resolve rate, defined as the proportion of generated patches that pass all relevant tests; all SWE-bench Lite results are reported with pass@1.

\textbf{Training Data Supplement: MathVista.} MathVista~\citep{lu2023mathvista} is not used as an evaluation benchmark in this paper, but only as a training supplement for CAT. We use 100 MathVista tasks for RFT training and 200 MathVista tasks for KTO trajectory sampling. The purpose of adding MathVista is to improve the execution agent's ability to handle GAIA-style multimodal attachments and visual reasoning inputs, rather than to expand the formal evaluation suite.

\subsection{Baselines}
\label{sec:baselines}

The baselines reported in this paper are organized following Table~\ref{tab:main_results}: general-purpose agents with static configuration, task-specialized agents with static configuration, pre-execution optimization, planner-worker orchestration, and configuration patching.

\textbf{General-Purpose Agents with Static Configuration.}

\textbf{Vanilla Agent.} A general-purpose static ReAct-style single agent~\citep{yao2022react}, equivalent to \name{} without \(T_{\text{reconfig}}\). It uses a fixed configuration \(\mathcal{C}_{\text{static}}\) throughout task execution, so sub-goal planning, strategy adjustment, toolbox selection, knowledge transfer, and context-management mode are not autonomously updated as runtime actions. This baseline measures the gain from incorporating self-reconfiguration into the execution agent's native action space. Vanilla Agent uses the basic context-management mode for FRAMES, GAIA, and xbench, and the truncation mode for SWE-bench Lite, as detailed in Appendix~\ref{sec:exp_details}.

\textbf{OpenHands.} OpenHands~\citep{wang2025openhands} is a general agent SDK that provides a composable and extensible agent runtime, tool interfaces, and execution environment, enabling models to solve complex tasks through tools such as shell commands, file editing, browsing, and code execution.

\textbf{Task-Specialized Agents with Static Configuration.}

\textbf{WebSailor.} WebSailor~\citep{li2025websailor} is a task-specialized single-agent method for deep research and high-uncertainty information retrieval. It uses tools such as \texttt{search} and \texttt{visit}, with retrieval-oriented prompting to support multi-step search, webpage access, and evidence integration.

\textbf{SWE-Agent.} SWE-Agent~\citep{yang2024swe} is a task-specialized agent for software engineering. It introduces the Agent-Computer Interface and provides navigation, file viewing, editing, and lint-aware execution mechanisms tailored to coding tasks, enabling bug localization and repair in real codebases. This baseline is mainly used on SWE-bench Lite.

\textbf{Co-Sight.} Co-Sight~\citep{zhang2025cosight} is a search/reasoning method. It improves the reliability of search-enhanced reasoning through conflict-aware meta-verification and structured facts, making it particularly suitable for deep research and complex information-integration tasks.

\textbf{SWE-Search.} SWE-Search~\citep{antoniades2024swe} is a software-engineering method. It systematically explores code-repair trajectories through Monte Carlo tree search and iterative refinement, thereby improving software issue resolution.

\textbf{Pre-execution Optimization.}

\textbf{AutoAgent.} AutoAgent~\citep{tang2025autoagent} is a pre-execution optimization method that aims to automatically construct or select a suitable agent or workflow configuration from natural-language requirements, reducing the manual cost of designing agent structures, tool combinations, and execution procedures.

\textbf{Planner-Worker Orchestration.}

\textbf{OWL.} OWL~\citep{hu2025owl} is a planner-worker orchestration baseline. It adopts a hierarchical multi-agent architecture in which a Planner decomposes the task, a Coordinator assigns subtasks, and specialized workers such as Web Agent, Document Agent, and Coding Agent execute them. Such methods cover heterogeneous tasks through division of labor, but task states must be repeatedly relayed, summarized, and compressed among the planner, coordinator, and workers, which can cause information loss and make final success or failure difficult to attribute clearly to planning decisions or worker execution behavior.

\textbf{Configuration Patching.}

\textbf{ReSum.} ReSum~\citep{wu2025resum} patches the context dimension. It mitigates context overload by periodically summarizing long-horizon search history into a compact state and continuing exploration from that state. This method mainly improves context management and does not jointly update the sub-goal, strategy, toolbox, knowledge, and context-management mode.

\textbf{OAgents.} OAgents~\citep{zhu2025oagents} patches the planning dimension. It uses an external LLM planner to periodically update the plan during execution, for example every five steps. This planning update is triggered by an external module at a fixed cadence, rather than being autonomously invoked by the execution agent within its own action space according to task progress.

\textbf{AgenTRIM.} AgenTRIM~\citep{betser2026agentrim} is a toolbox-level configuration patching method. During agent execution, it trims, filters, and validates the available tool set according to the current task state, exposing the agent to a smaller and more relevant tool space and thereby reducing the search burden and erroneous calls caused by irrelevant tools.

\textbf{Unified Evaluation Infrastructure.} To ensure fair comparison, we reuse the same underlying tool implementations as \name{} whenever permitted by the baseline framework. For baselines that depend on architecture-specific interfaces or execution logic, we retain their original tool invocation and execution procedures to avoid altering the method itself.

\subsection{Experimental Details}
\label{sec:exp_details}

\subsubsection{Execution Agent Runtime Configuration}

We use \texttt{Qwen3-8B} and \texttt{Qwen3-14B} as base models for the execution agent. The behavior at stage \(i\) is denoted by \(\pi_i\), which executes the task under the dynamic configuration \(\mathcal{C}_i=(q_i,\sigma_i,T_i,K_i,m_i)\). Here, \(q_i\) denotes the current-stage sub-goal, \(\sigma_i\) the execution strategy, \(T_i\) the available toolbox at the current stage, \(K_i\) the task-relevant knowledge transferred from previous stages, and \(m_i\) the context-management mode. The execution agent is implemented based on Vanilla Agent and follows the ReAct reasoning paradigm; its prompt template is provided in Sec.~\ref{sec:execution_agent_prompt}. Except for the Reconfiguration Tool, it shares the remaining settings with Vanilla Agent, as detailed in Appendix~\ref{sec:baselines}.

All benchmarks use the same inference parameters: \texttt{temperature=0.6} and \texttt{top\_p=0.95}. This setting balances stability and exploration, helping the agent avoid repetitive failure modes in long-horizon tasks while preserving the strategy exploration needed for adaptation.

We also set uniform and deliberately loose execution bounds for all tasks: a maximum iteration count of 200, a maximum reconfiguration count of 30, and a maximum context window length of 32,000 tokens, serving as safeguards against infinite loops and context overflow. Table~\ref{tab:safety_bounds} reports the trigger rates of the maximum-iteration and maximum-reconfiguration limits. The maximum iteration limit is never triggered on any benchmark, and the maximum reconfiguration limit is almost never triggered, with only a 1.9\% trigger rate on GAIA. These global thresholds therefore mainly serve as stability safeguards for extreme failures.

\begin{table}[H]
\caption{Trigger rates of maximum-iteration and maximum-reconfiguration limits across benchmarks.}
\label{tab:safety_bounds}
\centering
\begin{tabular}{lcc}
\toprule
\textbf{Benchmark} & \textbf{Max Iter (200)} & \textbf{Max Reconfig (30)} \\
\midrule
FRAMES & 0.0\% & 0.0\% \\
xbench & 0.0\% & 0.0\% \\
GAIA(WS) & 0.0\% & 0.0\% \\
GAIA & 0.0\% & 1.9\% \\
SWE-bench Lite & 0.0\% & 0.0\% \\
\bottomrule
\end{tabular}
\end{table}

\subsubsection{Configuration Generation Prompt}

Configuration generation uses the configuration-generation specification \(z_i=(\gamma_{\mathrm{cfg}}, Q, r_i, T_{\text{all}}, H_{\text{all}}^{(i)})\) defined in the Method section. Here, \(\gamma_{\mathrm{cfg}}\) specifies the output format and constraints for next-stage configuration generation; the full prompt is provided in Sec.~\ref{sec:config_generation_spec_prompt}.

\subsubsection{Model Training Configuration}

CAT aims to teach the execution agent runtime self-reconfiguration. Training proceeds in two stages: Stage I performs cold-start initialization via Rejection Sampling Fine-tuning (RFT), and Stage II performs reinforcement learning refinement via Kahneman-Tversky Optimization (KTO).

\paragraph{Training Data Construction.}
The training data are drawn from FRAMES and MathVista. FRAMES contains 824 questions, of which 200 are used as the test set and the remaining 624 as the training pool. MathVista is used only as a training supplement to improve the execution agent's ability to handle GAIA-style multimodal attachments and visual reasoning inputs, and is not used as a formal evaluation benchmark. For Stage I RFT training, we use 200 FRAMES tasks and 100 MathVista tasks. For Stage II KTO training, we use 424 FRAMES tasks and 200 MathVista tasks, totaling \(M'=624\). Teacher trajectory generation uses \texttt{Qwen3-235B-A22B-Thinking}. The RFT data are not unfiltered teacher rollouts; instead, teacher generation is combined with rejection sampling: after generating a trajectory for each task, we retain only trajectories judged successful by the official evaluation scripts. This yields 162 successful trajectories, with an average of 8.72 reconfigurations per trajectory. The teacher uses the same trajectory-generation settings as the student execution agent.

The teacher is used only for cold-start demonstrations and is not used at test time. We emphasize that zero-shot \name{} already consistently outperforms Vanilla Agent before CAT training, indicating that the main gains come from the unified self-reconfiguration mechanism itself.

\paragraph{LoRA Configuration.}
We use LoRA~\citep{hu2022lora} for parameter-efficient fine-tuning of the execution agent. The LoRA rank is set to \(r=16\), with scaling factor \(\alpha=32\). The base model parameters \(\theta_0\) remain frozen throughout training.

\paragraph{Stage I: RFT Training Details.}
RFT uses the AdamW optimizer with learning rate 8e-6, weight decay 0.08, \(\beta_1=0.9\), and \(\beta_2=0.95\). The per-device batch size is 1, gradient accumulation steps are 8, and the global batch size is 16. Training runs for 2 epochs with a cosine learning-rate scheduler, warmup ratio 0.15, and gradient clipping max norm 0.5. Training uses bfloat16 mixed precision and DeepSpeed ZeRO-2 on 2 NVIDIA A100 GPUs.

\paragraph{Stage II: KTO Training Details.}
KTO is initialized from the RFT checkpoint. For the \(M'=624\) training tasks, we sample \(K=4\) trajectories per task, including 424 FRAMES tasks and 200 MathVista tasks. Trajectory sampling uses \texttt{temperature=0.7} and \texttt{top\_p=0.9} to increase exploration diversity. Trajectory labels are assigned using the official evaluation metrics: success is labeled \(y=1\) and failure \(y=0\). We finally collect 2496 annotated trajectories, with a positive-to-negative sample ratio of approximately 0.83:1.

The KTO hyperparameters are set to \(\lambda_D=1.0\), \(\lambda_U=1.0\), and KL coefficient \(\beta=0.1\). The optimizer is AdamW with learning rate 3e-6, \(\beta_1=0.9\), \(\beta_2=0.95\), weight decay 0.01, and warmup ratio 0.1. The per-device batch size is 1, gradient accumulation steps are 4, and the global batch size is 4. Training runs for 1 epoch with gradient clipping max norm 1.0. The reference model is the frozen \(\theta^{\text{RFT}}\). Training uses bfloat16 mixed precision on 4 NVIDIA A100 GPUs.

\paragraph{Training Implementation Details.}
All training experiments are implemented with PyTorch 2.9.0 and Transformers 4.57.3. LoRA is implemented with PEFT 0.18.0, and the training framework is ms-swift 3.12.0. The RFT stage uses DeepSpeed 0.18.3 ZeRO-2 for distributed training on 2 GPUs; checkpoints are saved every 40 steps, with the most recent 3 retained. Training monitors training loss, validation loss, and token accuracy, and the final checkpoint is selected by the lowest eval\_loss on the validation set.

The KTO stage uses single-process mode on 4 GPUs. Checkpoints are saved every 300 steps, with the most recent 3 retained, and \texttt{load\_best\_model\_at\_end} is enabled. Training monitors training loss, validation loss, KTO reward margins, and positive/negative sample losses, and the final checkpoint is again selected by the lowest eval\_loss on the validation set. Both stages use bfloat16 mixed precision and gradient checkpointing to reduce memory usage.

\subsubsection{Core Tool Definitions}

\(T_{\text{reconfig}}\) and \(T_{\text{term}}\) are the two core management tools in \name{}. \(T_{\text{reconfig}}\) is invoked to trigger a configuration update when the current sub-goal is completed, the current strategy fails, the toolbox is insufficient, knowledge is missing, or the current context-management mode is no longer suitable for subsequent stages. It takes the current stage's execution summary and reconfiguration request as input and returns the configuration-generation specification \(z_i\), whose prompt component \(\gamma_{\text{cfg}}\) is inserted directly into the execution agent context. \(T_{\text{term}}\) is invoked only when the execution agent judges that the main task has been fully completed. In implementation, \(T_{\text{term}}\) is exposed as the \texttt{terminate} tool. The full tool-definition prompts for both tools are given in Sec.~\ref{sec:tool_definitions}.

\subsubsection{Global History Pool $H_{\text{all}}$}

The global history pool \(H_{\text{all}}\) stores execution summaries from all previous stages in string form. After each call to \(T_{\text{reconfig}}\), the current stage summary is added to \(H_{\text{all}}\). When generating the next-stage configuration, the execution agent selects or summarizes the key information from \(H_{\text{all}}\) to be passed into \(K_{i+1}\). Therefore, \(H_{\text{all}}\) serves as the main source of cross-stage knowledge transfer, rather than preserving the full raw interaction history within each stage.

\subsubsection{Global Tool Pool $T_{\text{all}}$}

The global tool pool \(T_{\text{all}}\) contains six tools: \texttt{visit}, \texttt{search}, \texttt{code\_interpreter}, \texttt{execute\_bash}, \texttt{str\_replace\_editor}, and \texttt{file\_analyzer}. At each stage, the execution agent can use only a subset \(T_i \subseteq T_{\text{all}}\) selected from the global tool pool. This keeps the action space compact during ordinary execution while allowing the reconfiguration stage to reselect tools according to subsequent needs.

\textbf{1. \texttt{visit} (Web Access Tool).} The tool definition prompt is shown in Sec.~\ref{sec:tool_definitions}. Its implementation follows WebSailor~\citep{li2025websailor}, using the Jina Reader API to access and parse webpage content and supporting both single-URL and batched URL access. It performs goal-directed content extraction based on the specified objective and returns only information and evidence relevant to the current goal, rather than lengthy raw webpage content. When needed, the content-extraction component uses the same backbone family as the execution agent.

\textbf{2. \texttt{search}.} The tool definition prompt is shown in Sec.~\ref{sec:tool_definitions}. Its implementation follows WebSailor~\citep{li2025websailor}, supporting batched query submission, automatic language detection, and concurrent request handling, and returns formatted results including title, link, summary, publication date, and source domain. For FRAMES, whose sources are mainly specific Wikipedia pages, we pre-download the relevant pages and deploy a BM25-based retrieval service; for the other tasks, we use SearxNG as the search backend.

\textbf{3. \texttt{code\_interpreter} (Python Code Interpreter).} The tool definition prompt is shown in Sec.~\ref{sec:tool_definitions}. It executes Python code in an isolated Jupyter Kernel environment, with each agent instance maintaining an independent kernel state and supporting cross-call variable persistence. Preinstalled libraries include NumPy, Pandas, Matplotlib, SciPy, and common scientific computing packages for data analysis, statistical modeling, visualization, and mathematical computation.

\textbf{4. \texttt{execute\_bash}.} The tool definition prompt is shown in Sec.~\ref{sec:tool_definitions}. Its implementation follows SWE-Agent~\citep{yang2024swe} and executes commands in an isolated shell session, supporting repository navigation, file-system operations, script execution, environment inspection, and test execution. The tool includes timeout control and output truncation to handle commands with long runtimes or excessively long outputs.

\textbf{5. \texttt{str\_replace\_editor}.} The tool definition prompt is shown in Sec.~\ref{sec:tool_definitions}. Its implementation follows SWE-Agent~\citep{yang2024swe} and supports operations such as \texttt{view}, \texttt{create}, \texttt{str\_replace}, \texttt{insert}, and \texttt{undo\_edit} for file inspection, precise string replacement, insertion, and edit rollback. This tool mainly serves patch editing in software engineering tasks.

\textbf{6. \texttt{file\_analyzer} (File Analysis Tool).} The tool definition prompt is shown in Sec.~\ref{sec:tool_definitions}. For text documents such as PDF, Word, Excel, PowerPoint, TXT, and Markdown files, it uses a document parser and goal-directed extraction to return task-relevant content; for image files such as JPG, PNG, GIF, BMP, and WebP, it uses \texttt{Qwen3-VL-30B-A3B-Instruct} for visual understanding. This tool is mainly used for multimodal GAIA tasks with attachments.

\subsubsection{Context-Management Modes}

In our experiments, \(m_i\) has two possible values: the basic mode and the truncation mode. They provide system-level support for stable execution across different task environments; \name{} treats the context-management mode as part of the runtime configuration, enabling the execution agent to manage it through self-reconfiguration within a unified configuration.

The basic mode applies to most task scenarios. When the visible context exceeds 80\% of the maximum context capacity, i.e., 25,600 tokens, the system clears historical interaction content while retaining the most recent 10 iterations.

The truncation mode is designed for scenarios involving very long individual tool calls or responses, such as software engineering or PDF reading and editing. In these settings, tool calls and returned text are often very large, and retaining them in the model context throughout execution can severely impair subsequent steps. This mode therefore applies a two-level truncation mechanism: first, single-message truncation performs head-preserving truncation on overly long non-current-iteration tool calls and responses, inserting explicit markers such as \texttt{... (truncated 1234 characters) ...}; second, observation-history truncation keeps only the 10 most recent complete tool responses and replaces earlier responses with concise placeholders such as \texttt{Old environment output: (42 lines omitted)}.

\subsubsection{Environment Interaction Configuration}

For SWE-bench Lite, each test instance runs in an independent Docker sandbox to ensure environment isolation and reproducibility. The execution agent primarily interacts with the repository through \texttt{execute\_bash} and \texttt{str\_replace\_editor}, using \texttt{code\_interpreter} when auxiliary computation or analysis is needed. Following common post-v1.0.1 SWE-Agent practice, \texttt{execute\_bash} is mainly used for repository navigation, command execution, and test verification, while \texttt{str\_replace\_editor} is mainly used for file inspection and patch editing.

The core tool requirements of SWE tasks are relatively stable, so reconfiguration does not necessarily manifest as frequent tool switching. On SWE-bench Lite, reconfiguration mainly updates stage-level sub-goals, repair and verification strategies, and cross-stage knowledge, enabling the execution agent to transition from bug localization to patch verification and to recover when failed tests or over-fixes expose issues. For non-SWE tasks, the above tools interact directly with the local evaluation environment.

\subsubsection{Model Evaluation}

All benchmark results are computed using official evaluation metrics and scripts whenever possible to ensure comparability with existing baselines. FRAMES, xbench, and GAIA use LLM-as-Judge to determine whether model outputs match the ground-truth answers. The judge model is \texttt{Qwen2.5-72B-Instruct}, with inference parameters \texttt{temperature=0.0} and \texttt{top\_p=1.0}.

To assess the reliability of LLM-as-Judge, we manually verify \texttt{Qwen3-14B} \name{} outputs, as summarized in Table~\ref{tab:judge_agreement}. For FRAMES, we randomly sample 100 samples, covering 50\% of the test set, and obtain a human-judge agreement rate of 97.0\%. For GAIA, we randomly sample 83 samples, also covering 50\%, with an agreement rate of 96.4\%. For xbench, we verify all 100 samples, covering 100\%, with an agreement rate of 98.0\%. We also manually inspect all cases where automatic judgments disagree with human annotations. The results show that LLM-as-Judge is highly consistent with human judgments and can support the main performance comparisons in this paper.

\begin{table}[H]
\caption{Human verification sample sizes and agreement rates for \texttt{Qwen3-14B} \name{} outputs.}
\label{tab:judge_agreement}
\centering
\begin{tabular}{lccc}
\toprule
\textbf{Benchmark} & \textbf{Samples} & \textbf{Sampling Ratio} & \textbf{Agreement Rate} \\
\midrule
FRAMES & 100 random samples & 50\% & 97.0\% \\
GAIA & 83 random samples & 50\% & 96.4\% \\
xbench & 100 samples & 100\% & 98.0\% \\
\bottomrule
\end{tabular}
\end{table}

SWE-bench Lite uses resolve rate as the evaluation metric, defined as the proportion of generated patches that pass all relevant tests. All SWE-bench Lite results are reported with pass@1.

\subsubsection{Computational Resources}

All experiments are run on a high-performance computing node equipped with 16 NVIDIA A100 GPUs, each with 40 GB of memory. RFT training uses 2 NVIDIA A100 GPUs, KTO training uses 4 NVIDIA A100 GPUs, and evaluation uses up to 16 NVIDIA A100 GPUs depending on the benchmark and model scale. The evaluation environment uses PyTorch 2.4.0, CUDA 12.1, and NCCL 2.20.5. The training implementation uses PyTorch 2.9.0, Transformers 4.57.3, PEFT 0.18.0, ms-swift 3.12.0, and DeepSpeed 0.18.3; these version differences reflect the actual deployments of the evaluation and training environments.

\subsubsection{Existing Assets and Licenses}

This work uses existing benchmarks, models, and tool implementations, including FRAMES, xbench, GAIA, SWE-bench Lite, MathVista, Qwen-series models, \texttt{MiniMax-M2.5}, SWE-Agent, and WebSailor. We cite the original papers or official resources for these assets and follow their official licenses and terms of use.

\subsection{Limitations}
\label{sec:limitations}

First, this work focuses on whether an agent can internalize the configuration of its sub-goal, strategy, toolbox, and context management as learnable actions under a fixed global tool pool.
Thus, \name{} can dynamically select tools from \(T_{\text{all}}\) at runtime, but it does not create new tools for unseen tasks.
Dynamic tool creation and tool synthesis are orthogonal and valuable directions for future work.
Second, since self-reconfiguration introduces additional model calls, deploying \name{} in real-time systems may incur extra latency.

\subsection{Prompt Templates}
\label{sec:prompt_templates}

\subsubsection{Trajectory Capability Judge Prompt}
\label{sec:trajectory_capability_judge_prompt}

For the trajectory analysis in Table~\ref{tab:adaptive_capabilities}, we concatenate the realized stage-level sub-goals of each GAIA trajectory and use an LLM-as-Judge to classify observable adaptive behaviors. The judge model is \texttt{Qwen2.5-72B-Instruct}, with inference parameters \texttt{temperature=0.0} and \texttt{top\_p=1.0}.

\begin{promptbox}{Trajectory Capability Judge Prompt}
\tiny
\begin{verbatim}
You are a rigorous evaluator of AI-agent sub-goal trajectories.

Your task is to classify observable adaptive behaviors from a concatenated
sequence of realized stage-level sub-goals.

Important rules:
- Use only the provided concatenated stage-level sub-goals.
- Treat each listed sub-goal as a realized stage objective.
- Do not infer hidden reasoning, execution details, tool outputs, failures,
  or answer correctness.
- Do not use the task outcome or final answer correctness as evidence.
- If a behavior is not explicitly observable from the sub-goal sequence,
  mark it as absent.
- A capability is present only when there is concrete evidence in the realized
  sub-goal sequence.
- Return strict JSON only. Do not include markdown or explanations outside JSON.

Capability definitions:
- long_horizon_planning: The realized sub-goal sequence forms a
  dependency-ordered multi-step plan. Count this when the stages decompose the
  task into ordered subtasks, such as identify -> retrieve -> compute ->
  verify -> answer.
- self_refinement: A later sub-goal narrows, constrains, or makes a previous
  coarse sub-goal more specific.
- self_correction: A later sub-goal explicitly indicates recovery from an
  error, failed attempt, ambiguity, wrong assumption, missing information, or
  insufficient result. Do not infer correction unless the sub-goal text itself
  signals it.
- verification: A later sub-goal explicitly verifies, confirms, validates,
  cross-checks, reconciles, or audits a candidate fact, calculation, source,
  or answer.
- strategy_adaptation: A later sub-goal explicitly changes method, source type,
  tool use, search strategy, computation strategy, or context strategy in
  response to a limitation. Do not infer adaptation unless the sub-goal text
  itself signals a changed approach.

Input:
Task ID: {{task_id}}
Run label: {{run_label}}
Question: {{question}}
Concatenated stage-level sub-goals:
{{concatenated_stage_level_subgoals}}

Return exactly this JSON structure:
{
  "task_id": "{{task_id}}",
  "run_label": "{{run_label}}",
  "capabilities": {
    "long_horizon_planning": {
      "present": true or false,
      "evidence": "brief evidence from the sub-goal sequence"
    },
    "self_refinement": {
      "present": true or false,
      "evidence": "brief evidence from the sub-goal sequence"
    },
    "self_correction": {
      "present": true or false,
      "evidence": "brief evidence from the sub-goal sequence"
    },
    "verification": {
      "present": true or false,
      "evidence": "brief evidence from the sub-goal sequence"
    },
    "strategy_adaptation": {
      "present": true or false,
      "evidence": "brief evidence from the sub-goal sequence"
    }
  },
  "overall_notes": "one brief sentence based only on the sub-goal sequence"
}
\end{verbatim}
\end{promptbox}

\clearpage
\subsubsection{Execution Agent ($\pi_i$) Prompt Template}
\label{sec:execution_agent_prompt}

The execution agent uses a three-part prompt structure: the SYSTEM section provides configuration (main task, sub-goal, strategy, toolbox, knowledge, context-management mode), the USER section contains tool definitions and ReAct format instructions, and the ASSISTANT section records execution history.

\begin{promptbox}{Execution Agent System Prompt}
\fontsize{5pt}{5.5pt}\selectfont
\begin{verbatim}
=== SYSTEM ===
## Role and Core Objective
You are a professional Execution Agent. Your core objective is to
efficiently complete the specified task using the given main task, sub-goal,
and available toolset.

## Input Information
You will receive the following information, please analyze carefully:

1. **Main Task `<main_task>`**: This is the ultimate goal to be completed.
   <main_task>
   {MAIN_TASK_CONTENT}
   </main_task>

2. **Current Sub_goal `<sub_goal>`**: This is the specific task you need to
   focus on completing now.
   <sub_goal>
   {SUB_GOAL_CONTENT}
   </sub_goal>

3. **Execution Strategy `<strategy>`**: The execution methodology for the
   current sub_goal.
   <strategy>
   {EXECUTION_STRATEGY}
   </strategy>

4. **Toolbox `<toolbox>`**: This is the list of tools available for the
   current sub_goal.
   <toolbox>
   {TOOLBOX_LIST}
   </toolbox>

5. **Knowledge Information `<knowledge>`**: This is the background information
   and constraints related to the task.
   <knowledge>
   {KNOWLEDGE_CONTENT}
   </knowledge>

6. **Context-Management Mode `<context_management_mode>`**: This is the
   current runtime mode that controls how the agent maintains, compresses,
   truncates, and organizes the visible context.
   <context_management_mode>
   {CONTEXT_MANAGEMENT_MODE}
   </context_management_mode>

## Tool Usage Guidelines
- **Precision**: Select the most appropriate tools based on the current
  sub-goal
- **Efficiency**: Prioritize tools that can directly solve the problem
- **Completeness**: Ensure the task execution process is complete and
  logically clear
- **Context Awareness**: Respect the current context-management mode when
  reading history and processing long tool outputs.

## Agent Management Tools
**Reconfiguration Tool** - Use when the Execution Agent needs configuration updates:
- Current sub-task completed but main task not finished → update sub_goal
- Current sub-task execution failed → try new sub-task approach
- Current toolbox insufficient → update toolbox
- Knowledge lacks necessary information → update knowledge
- Execution strategy no longer fits → update execution_strategy
- Current context-management mode no longer fits → update
  context_management_mode

**Termination Tool** - Use only when:
- Main task is completely finished (not just sub-goal)
- All objectives have been achieved and no further work is needed

## Execution Requirements
1. **CRITICAL - Focus ONLY on Sub_goal**: Your ONLY objective is to complete
   the current sub_goal. DO NOT attempt to complete the entire main task.
2. **Main Task Usage**: Use the main task information ONLY to understand the
   broader context and make informed decisions about the current sub_goal.
3. **Tool Selection**: Choose appropriate tools from the toolbox
   to complete the sub_goal.
4. **Result Reporting**: Use the termination tool only when the MAIN TASK is
   completely finished. Use the reconfiguration tool when needing to update Execution Agent
   configuration.
5. **Runtime Configuration**: Treat `<context_management_mode>` as an explicit
   part of the current runtime configuration, not as hidden implementation
   detail.
6. **Error Handling**: When encountering problems, use the reconfiguration tool to
   modify agent configuration rather than simply reporting errors.
\end{verbatim}
\end{promptbox}

\clearpage
\subsubsection{ReAct Execution Format}
\label{sec:react_format}

\begin{promptbox}{ReAct Format Specification}
\tiny
\begin{verbatim}
The execution agent operates within a structured prompt format:

USER_PROMPT_PREFIX:
"A conversation between User and Assistant. The user asks a question, and the
assistant solves it by calling one or more of the following tools."

[Tool definitions inserted here]

USER_PROMPT_SUFFIX:
"The assistant starts with one or more cycles of (thinking about which tool to
use -> performing tool call -> waiting for tool response), and ends by calling
the `terminate` tool (implementation name of T_term) with a final summary and
completion status. The thinking
processes, tool calls, and tool responses are enclosed within their tags."

## Execution Cycle Structure

<think> [reasoning about current situation and next action] </think>
<tool_call>
{"name": "tool_name", "arguments": {"arg1": "value1", "arg2": "value2"}}
</tool_call>
<tool_response>
[Environment's response]
</tool_response>
[Multiple cycles continue...]
<think> [final reasoning] </think>
<tool_call>
{"name": "terminate", "arguments": {
  "task_completion_status": "complete",
  "final_result": "result summary",
  "execution_summary": {
    "detailed_execution": ["step1", "step2"],
    "tools_used": ["tool1", "tool2"]
  }
}}
</tool_call>
\end{verbatim}
\end{promptbox}

\clearpage
\subsubsection{Configuration-Generation Specification ($\gamma_{\text{cfg}}$) Prompt Template}
\label{sec:config_generation_spec_prompt}

\begin{promptbox}{Configuration-Generation Specification}
\fontsize{5pt}{5.5pt}\selectfont
\begin{verbatim}
## Role and Core Objective
You are a configuration generator for a ReAct-based Execution Agent. Your core
objective is to generate the next-stage runtime configuration for the Execution Agent
executing a complex main task. You must:
1. Mentally plan the overall task execution to understand the big picture
2. Identify the immediate next step based on current progress
3. Generate appropriate configuration (sub_goal, strategy, toolbox, knowledge,
   context_management_mode)

## Input Information
You will receive the following parts:
1. **Main Task `<main_task>`**: The final goal the Execution Agent needs to accomplish.
   <main_task>
   {MAIN_TASK_CONTENT}
   </main_task>
2. **Available Tools `<available_tools>`**: Complete list of tools available
   to the Execution Agent during the entire task lifecycle.
   <available_tools>
   {ALL_AVAILABLE_TOOLS}
   </available_tools>

3. **Context-Management Modes `<context_management_modes>`**: Complete list of
   context-management modes available for the next execution stage. Select
   exactly one mode from this list as `context_management_mode`.
   <context_management_modes>
   [
     {
       "name": "basic",
       "description": "General mode for ordinary reasoning, web search, and
         tool-use stages. When the visible context becomes too long, historical
         interaction content is cleared while recent iterations are retained."
     },
     {
       "name": "truncation",
       "description": "Mode for stages likely to involve very long tool calls
         or tool responses, such as software engineering, PDF reading,
         document analysis, or file editing. Long non-current tool messages
         and older observations may be truncated or replaced with placeholders."
     }
   ]
   </context_management_modes>

4. **Execution History `<execution_history>`**: Summarized record of all
   previous steps. Can be empty ("NONE").
   <execution_history>
   {EXECUTION_HISTORY}
   </execution_history>

5. **Update Requirement `<update_requirement>`**: Update suggestions from
   the current Execution Agent. Can be empty ("NONE").
   <update_requirement>
   {UPDATE_REQUIREMENT}
   </update_requirement>

## Configuration Generation Process
### Step 1: Determine Next Sub-Goal
Carefully consider the proposed sub-goal in `<update_requirement>` and adopt
it whenever possible. Only re-plan a new sub-goal if the proposal is entirely
irrelevant to the task.
### Step 2: Generate Other Configuration Components
Based on the determined sub-goal and other requirements specified in
`<update_requirement>`, generate the following elements:
**2.1 Execution Strategy**
For **execution_strategy**, design a heuristic algorithm specifically for
the next sub_goal:
1. **Define Persona**: Assign an expert persona (e.g., "Senior Software
   Engineer", "Data Analyst").
2. **Create a Step-by-Step Plan**: Simulate the thinking process to
   accomplish the next sub_goal.
3. **Identify Decision Points**: Mark key decision points in the plan.
4. **Link to Tools**: Specify which concrete tool should be used in each step.
**2.2 Toolbox Selection**
Select tools for the **toolbox**:
- **CRITICAL**: The toolbox must include at least two tools specifically
  needed for the NEXT sub_goal ONLY.
- **VERY IMPORTANT**: Focus EXCLUSIVELY on tools required to complete the
  next sub_goal.
- **IMPORTANT**: The toolbox must be a **strict subset** of
  `<available_tools>`.
**2.3 Inter-Agent Knowledge**
Determine the `inter_agent_knowledge` field based on `<execution_history>`:
1. **Initial Step**: If `<execution_history>` is `NONE` or empty,
   `inter_agent_knowledge` must be an empty string `""`.
2. **Use ALL**: If history is concise and relevant, use the string `"ALL"`.
3. **Summarize**: If history is long or contains exploratory steps, extract
   a concise summary.
**2.4 Context-Management Mode**
Select the `context_management_mode` field from `<context_management_modes>`
based on the next sub_goal, expected tool outputs, and context requirements:
1. Use `"basic"` for ordinary reasoning, web search, and tool-use stages where
   tool outputs are expected to be moderate.
2. Use `"truncation"` when the next stage is likely to involve very long tool
   calls or tool responses, such as software engineering, PDF/document analysis,
   repository editing, or other context-heavy interactions.
## Output Format
Your final output **must** be a strict JSON object:
{
  "next_sub_goal": "Detailed description of the next step to execute",
  "execution_strategy": "As a [expert persona], I will follow these
    steps to accomplish the NEXT sub_goal: 1. [first step and tool]
    2. [second step and tool] 3. [key decision points]...",
  "toolbox": ["ToolName1", "ToolName2", "ToolName3"],
  "inter_agent_knowledge": "One of three forms: summary, 'ALL', or ''",
  "context_management_mode": "One of: basic, truncation"
}
\end{verbatim}
\end{promptbox}

\clearpage
\subsubsection{Tool Definitions}
\label{sec:tool_definitions}

The reconfiguration tool arguments correspond to the formal definitions in Sec.~\ref{sec:Method} as follows: \texttt{new\_sub\_goal} corresponds to $q_{i+1}^{\text{prop}}$, \texttt{update\_reason} to $\rho_i$, and \texttt{additional\_details} to $\delta_i$ in the reconfiguration request $r_i = (q_{i+1}^{\text{prop}}, \rho_i, \delta_i)$. Additionally, \texttt{execution\_summary} corresponds to the execution summary $H_i$. The tool returns the configuration-generation specification \(z_i\), whose prompt component is \(\gamma_{\text{cfg}}\), and this specification is inserted directly into the execution agent's context.

\begin{promptbox}{Reconfiguration Tool Definition}
\tiny
\begin{verbatim}
{
  "name": "reconfigure",
  "description": "Return a configuration-generation specification for the
    next execution stage. Use when the current sub-task is completed but the
    main task is not, when current sub-task execution fails, when the current
    toolbox is insufficient, when knowledge lacks necessary information or
    contains irrelevant information, when the execution strategy no longer
    fits the task requirements, or when the context-management mode should
    change for subsequent execution.",
  "arguments": {
    "type": "object",
    "properties": {
      "execution_summary": {
        "type": "string",
        "description": "Step-by-step detailed summary of current Execution Agent's
          task execution process"
      },
      "update_reason": {
        "type": "string",
        "description": "Why this update is needed"
      },
      "new_sub_goal": {
        "type": "string",
        "description": "The new sub_goal that the Execution Agent needs to
          execute. If the current task is complete, set to 'Task completed,
          use terminate tool next'"
      },
      "additional_details": {
        "type": "object",
        "properties": {
          "toolbox_requirements": {
            "type": "string",
            "description": "Requirements for the new toolbox based on
              current limitations"
          },
          "knowledge_requirements": {
            "type": "string",
            "description": "Requirements for the new knowledge based on
              missing information"
          },
          "execution_strategy_requirements": {
            "type": "string",
            "description": "Requirements for the new execution strategy based on
              current limitations"
          },
          "context_management_requirements": {
            "type": "string",
            "description": "Requirements for the next context-management mode
              based on expected tool outputs, history length, or context-heavy
              interactions"
          }
        },
        "description": "Required adjustments for generating the next
          configuration"
      }
    },
    "required": ["execution_summary", "update_reason", "new_sub_goal"]
  }
}
\end{verbatim}
\end{promptbox}

\begin{promptbox}{Terminate Tool Definition}
\tiny
\begin{verbatim}
{
  "name": "terminate",
  "description": "Signal the completion of the main task and end the entire
    task execution. Use only when the current Execution Agent has completely finished
    the main task.",
  "arguments": {
    "type": "object",
    "properties": {
      "task_completion_status": {
        "type": "string",
        "enum": ["complete", "partial", "incomplete"],
        "description": "Main task completion status"
      },
      "final_result": {
        "type": "string",
        "description": "Final result of the main task execution"
      },
      "execution_summary": {
        "type": "object",
        "properties": {
          "detailed_execution": {
            "type": "array",
            "items": {"type": "string"},
            "description": "Detailed step-by-step execution process including
              steps taken, key decisions made, challenges encountered and
              how resolved, blocking factors, progress updates, and critical
              insights."
          },
          "tools_used": {
            "type": "array",
            "items": {"type": "string"},
            "description": "List of tools used during the task"
          },
          "key_achievements": {
            "type": "array",
            "items": {"type": "string"},
            "description": "Major achievements and milestones reached"
          }
        },
        "required": ["detailed_execution", "tools_used"],
        "description": "Comprehensive summary of the task execution process"
      },
      "results_details": {
        "type": "object",
        "properties": {
          "deliverables": {
            "type": "array",
            "items": {"type": "string"},
            "description": "Concrete deliverables produced"
          },
          "evidence": {
            "type": "array",
            "items": {"type": "string"},
            "description": "Evidence supporting the results"
          },
          "confidence_level": {
            "type": "string",
            "enum": ["High", "Medium", "Low"],
            "description": "Confidence level in the results"
          }
        },
        "description": "Detailed information about the results achieved"
      },
      "issues_encountered": {
        "type": "object",
        "properties": {
          "blocking_issues": {
            "type": "array",
            "items": {"type": "string"},
            "description": "Issues that prevented full completion"
          },
          "workarounds_used": {
            "type": "array",
            "items": {"type": "string"},
            "description": "Workarounds employed to overcome issues"
          },
          "unresolved_issues": {
            "type": "array",
            "items": {"type": "string"},
            "description": "Issues that remain unresolved"
          }
        },
        "description": "Issues encountered during execution"
      }
    },
    "required": ["task_completion_status", "final_result",
                 "execution_summary"]
  }
}
\end{verbatim}
\end{promptbox}

\clearpage
\begin{promptbox}{Visit Tool Definition}
\tiny
\begin{verbatim}
{
  "name": "visit",
  "description": "Visit webpage(s) and return the summary of the content.",
  "arguments": {
    "type": "object",
    "properties": {
      "url": {
        "type": ["string", "array"],
        "items": {"type": "string"},
        "minItems": 1,
        "description": "The URL(s) of the webpage(s) to visit."
      },
      "goal": {
        "type": "string",
        "description": "The goal of the visit for webpage(s)."
      }
    },
    "required": ["url", "goal"]
  }
}
\end{verbatim}
\end{promptbox}

\begin{promptbox}{Search Tool Definition}
\tiny
\begin{verbatim}
{
  "name": "search",
  "description": "Performs batched web searches via a search backend.",
  "arguments": {
    "type": "object",
    "properties": {
      "query": {
        "type": "array",
        "items": {"type": "string"},
        "description": "Array of query strings."
      }
    },
    "required": ["query"]
  }
}
\end{verbatim}
\end{promptbox}

\begin{promptbox}{File Analyzer Tool Definition}
\tiny
\begin{verbatim}
{
  "name": "file_analyzer",
  "description": "Intelligently analyze and extract information from
    various file types including documents and images.",
  "arguments": {
    "type": "object",
    "properties": {
      "file_path": {
        "type": "string",
        "description": "The absolute or relative path to the file."
      },
      "goal": {
        "type": "string",
        "description": "A clear, specific question or goal describing what
          information you want to extract from the file."
      }
    },
    "required": ["file_path", "goal"]
  }
}
\end{verbatim}
\end{promptbox}

\clearpage
\begin{promptbox}{Bash Tool Definition}
\tiny
\begin{verbatim}
{
  "name": "execute_bash",
  "description": "Execute bash commands with safety checks. Supports timeout,
    working directory specification, and output truncation. Use absolute
    paths when possible. Dangerous commands are blocked.",
  "arguments": {
    "type": "object",
    "properties": {
      "command": {
        "type": "string",
        "description": "The bash command to execute (use '&&' to chain
          commands)."
      },
      "timeout": {
        "type": "integer",
        "description": "Timeout in seconds (default 120)."
      },
      "cwd": {
        "type": "string",
        "description": "Optional working directory (absolute path
          recommended)."
      }
    },
    "required": ["command"]
  }
}
\end{verbatim}
\end{promptbox}

\begin{promptbox}{Code Interpreter Tool Definition}
\tiny
\begin{verbatim}
{
  "name": "code_interpreter",
  "description": "Python code sandbox, which can be used to execute Python
    code.",
  "arguments": {
    "type": "object",
    "properties": {
      "code": {
        "type": "string",
        "description": "The python code."
      }
    },
    "required": ["code"]
  }
}
\end{verbatim}
\end{promptbox}

\clearpage
\begin{promptbox}{String Replace Editor Tool Definition}
\tiny
\begin{verbatim}
{
  "name": "str_replace_editor",
  "description": "Advanced file editor with intelligent
    features. Commands: view (display file/directory contents with optional
    line range), create (create a new file, cannot overwrite existing files),
    str_replace (replace text, old_str must match EXACTLY and be UNIQUE),
    insert (insert text at a specific line number), undo_edit (revert the
    last edit operation). Key Features: automatic tab expansion, shows edited
    code snippet with context after changes, smart window expansion to include
    complete functions/classes, integrated linter to detect syntax errors
    immediately, edit history with undo support, detailed error messages to
    help diagnose issues. IMPORTANT for str_replace: old_str must match the
    file content EXACTLY (character by character), old_str must be UNIQUE in
    the file (appears only once), always use 'view' first to get the exact
    text to replace, include enough context to make old_str unique, pay
    attention to whitespace and indentation.",
  "arguments": {
    "type": "object",
    "properties": {
      "command": {
        "type": "string",
        "enum": ["view", "create", "str_replace", "insert", "undo_edit"],
        "description": "Operation to perform."
      },
      "path": {
        "type": "string",
        "description": "Absolute file or directory path."
      },
      "file_text": {
        "type": "string",
        "description": "Required for 'create': new file content."
      },
      "old_str": {
        "type": "string",
        "description": "Required for 'str_replace': The exact text to replace.
          CRITICAL: Must match the file content EXACTLY (character-by-
          character, including all whitespace). Must appear only ONCE in the
          file. BEST PRACTICE: First use 'view' to see the exact content,
          then copy it here verbatim."
      },
      "new_str": {
        "type": "string",
        "description": "New text for 'str_replace' or 'insert'."
      },
      "insert_line": {
        "type": "integer",
        "description": "Required for 'insert': line number to insert at
          (0-indexed)."
      },
      "view_range": {
        "type": "array",
        "items": {"type": "integer"},
        "description": "Optional for 'view': [start, end], 1-indexed, -1 for
          end of file."
      }
    },
    "required": ["command", "path"]
  }
}
\end{verbatim}
\end{promptbox}

\clearpage
\subsubsection{Execution History Format}
\label{sec:execution_history_format}

\begin{promptbox}{Execution History Structure}
\tiny
\begin{verbatim}
<execution_history>
Iteration 1:
Sub-goal: [Description of the sub-goal for stage 0]
Summary: [Execution summary including completed work, encountered issues,
  key findings, and decisions made during stage 0]

Iteration 2:
Sub-goal: [Description of the sub-goal for stage 1]
Summary: [Execution summary for stage 1]

...

Iteration N:
Sub-goal: [Description of the sub-goal for stage N-1]
Summary: [Execution summary for stage N-1]
</execution_history>
\end{verbatim}
\end{promptbox}

\end{document}